\documentclass[10pt,journal,compsoc]{IEEEtran}
\ifCLASSOPTIONcompsoc
  \usepackage[nocompress]{cite}
\else
  \usepackage{cite}
\fi
\ifCLASSINFOpdf
\else
\fi
\usepackage{graphicx}
\usepackage{algorithm}
\usepackage{algorithmic}
\usepackage{booktabs}
\usepackage{subcaption}
\usepackage{amsmath,amssymb} 
\usepackage{comment}
\usepackage{bm}

\begin{document}
%
\title{Optimizing Class Distribution in Memory for \\ Multi-Label Online Continual Learning}
%
%
%
%

\author{Yan-Shuo Liang 
        and~Wu-Jun Li,~\IEEEmembership{Member,~IEEE,}
\IEEEcompsocitemizethanks{\IEEEcompsocthanksitem The authors are with the National Key Laboratory for Novel Software
Technology, Collaborative Innovation Center of Novel Software Technology
and Industrialization, Department of Computer Science and Technology, Nanjing University, Nanjing 210023, China.\protect\\
E-mail: liangys@smail.nju.edu.cn; liwujun@nju.edu.cn
}
\thanks{Manuscript received xxx, xxx; revised xxx, xxx.}}

%
%

\markboth{Journal of \LaTeX\ Class Files,~Vol.~14, No.~8, August~2015}%
{Shell \MakeLowercase{\textit{et al.}}: Bare Demo of IEEEtran.cls for Computer Society Journals}
%



\IEEEtitleabstractindextext{%
\begin{abstract}
  Online continual learning, especially when task identities and task boundaries are unavailable, is a challenging continual learning setting. One representative kind of methods for online continual learning is replay-based methods, in which a replay buffer called memory is maintained to keep a small part of past samples for overcoming catastrophic forgetting. When tackling with online continual learning, most existing replay-based methods focus on single-label problems in which each sample in the data stream has only one label. But multi-label problems may also happen in the online continual learning setting in which each sample may have more than one label. In the online setting with multi-label samples, the class distribution in data stream is typically highly imbalanced, and it is challenging to control class distribution in memory since changing the number of samples belonging to one class may affect the number of samples belonging to other classes. But class distribution in memory is critical for replay-based memory to get good performance, especially when the class distribution in data stream is highly imbalanced. In this paper, we propose a simple but effective method, called \underline{o}ptimizing \underline{c}lass \underline{d}istribution in \underline{m}emory~(OCDM), for multi-label online continual learning. OCDM formulates the memory update mechanism as an optimization problem and updates the memory by solving this problem. Experiments on two widely used multi-label datasets show that OCDM can control the class distribution in memory well and can outperform other state-of-the-art methods.
\end{abstract}

}

\maketitle

\IEEEdisplaynontitleabstractindextext

%
\IEEEpeerreviewmaketitle

\IEEEraisesectionheading{\section{Introduction}\label{sec:introduction}}
The ability to learn multiple tasks continuously, which is usually called continual learning, is essential for humans and other animals to live in a real environment~\cite{parisi2019continual}. Agents with continual learning ability can learn new tasks without retraining on old tasks. However, deep neural networks typically lack this ability. Specifically, when learning on the data with new distribution, the neural network may forget the old knowledge it has learned, so its performance on the old data distribution will drop significantly. This problem is often called catastrophic forgetting~\cite{goodfellow2013empirical,zheng2021evolving}.

Typically, in continual learning, the agent cannot obtain samples of the old task, or can only obtain a small part of the samples of the old task when learning a new task~\cite{parisi2019continual}. All the continual settings need to meet this requirement. One of the most popular settings is offline continual learning~\cite{hou2019learning, rajasegaran2020itaml, zhu2021prototype}, in which the agent receives all data from a single task at once. To achieve satisfactory results, methods based on this setting typically traverse the data of the current task multiple times. In addition to offline continual learning, some studies consider online continual learning~\cite{lopez2017gradient} where tasks are organized into a non-stationary data stream. The agent can only receive a mini-batch of task samples from the data stream, and the data of each task can only be traversed once. Furthermore, some works consider a more challenging online continual setting in which the agent is not provided with task boundaries or task identities throughout training~\cite{aljundi2019task, aljundi2019gradient, chrysakis2020online}. This setting is more in line with real-world scenarios, but also more challenging~\cite{DBLP:conf/iclr/TangM21}. Without task identities and boundaries, the agent does not know when the old task ends and the new task begins. This impedes the application of many offline and online continual learning methods that rely on task boundaries to determine when to learn a new task and when to consolidate prior task knowledge~\cite{aljundi2019task}. 

Several methods are proposed to address continual learning, including regularization-based methods~\cite{kirkpatrick2017overcoming, zenke2017continual, ahn2019uncertainty, shi2021continual}, expansion-based methods~\cite{yoon2017lifelong, li2019learn, hung2019compacting}, and replay-based methods~\cite{de2019continual, lopez2017gradient, guo2019improved}. In online continual learning, especially the task identities and task boundaries are not provided, replay-based methods consistently exhibit competitive performance~\cite{kim2020imbalanced,aljundi2019online}. These methods overcome catastrophic forgetting by storing a small amount of previous data in a memory buffer. In online continual learning, the challenges of these methods are how to employ saved limited samples to better overcome catastrophic forgetting and how to update the memory in the absence of task identities and boundaries. Although many methods are proposed for these problems, the majority of them only take into account the single-label setting~\cite{aljundi2019online,aljundi2019gradient,chrysakis2020online}, where each sample in the data stream has just one label. However, multi-label problems may also occur in the online continual learning, where each sample may have more than one label. For example, our humans learn a variety of things as we grow in the real environment without having clear task identities or task boundaries. We learn to recognize new items when we see them, like cats or dogs. However, the images we see might have more than one label because different classes of items might come into view at the same time. Learning from these images will be a multi-label problem. Therefore, it is necessary to consider online continual learning with multi-label samples.

In a data stream consisting of multi-label samples, the distribution for different classes is always imbalanced. This is different from the previously discussed single-label online continual learning setting, in which the data stream can be either balanced or imbalanced~\cite{aljundi2019online,chrysakis2020online}. In the data stream with imbalanced class distribution, the class distribution in memory is also at risk of becoming imbalanced, and existing research~\cite{aljundi2019gradient,chrysakis2020online} has demonstrated that imbalanced class distribution in memory is harmful to the agent's performance in continual learning. Furthermore, controlling class distribution in memory is more challenging than that in single-label problems since changing the number of samples belonging to one class may affect the number of samples belonging to other classes~(See Section~\ref{selection strategy} for a more specific example). Therefore, memory updating in multi-label setting is more challenging than single-label setting. When dealing with online continual learning without task identities and task boundaries, partition reservoir sampling~(PRS)~\cite{kim2020imbalanced} takes into account the multi-label problem and can control class distribution in memory. However, as demonstrated by the experimental results in this work, this method suffers from slow speed due to its complicated procedure. But the speed is important since the agent needs to react quickly to the changes in the real environment.

In this paper, we propose a simple but effective memory updating method, called \underline{o}ptimizing \underline{c}lass \underline{d}istribution in \underline{m}emory~(OCDM), for multi-label online continual learning setting when the task identities and task boundaries are unavailable. 
The main contributions of OCDM are outlined as follows:
\begin{itemize}
    \item OCDM formulates the memory update process for online continual learning into a selection problem and proposes an optimization problem to represent it.
    \item OCDM proposes a greedy algorithm to solve the above optimization problem, which can control the class distribution in memory well when the data stream consists of multi-label samples.
    \item Experiments on two widely used multi-label datasets MSCOCO and NUS-WIDE show that OCDM outperforms other state-of-the-art methods including PRS in terms of accuracy, and its speed is also much faster than PRS.
\end{itemize}

\section{Related Work}
Three types of methods are proposed to tackle continual learning problems, including regularization-based methods~\cite{kirkpatrick2017overcoming, zenke2017continual, ahn2019uncertainty, shi2021continual}, expansion-based methods~\cite{yoon2017lifelong, li2019learn, hung2019compacting}, and replay-based methods~\cite{de2019continual, lopez2017gradient, guo2019improved}. Regularization-based methods add a penalty to the loss function and minimize penalty loss with new task loss together. Expansion-based methods expand the network's architecture when learning a new task. Replay-based methods maintain a memory to keep a small part of past samples. In this work, we consider replay-based methods which have shown consistently competitive performance in the online continual learning setting, especially when the task identities and task boundaries are not provided to the agent~\cite{kim2020imbalanced, aljundi2019online}.

In the online continual learning setting, replay-based methods usually consider two problems, including how to employ limited save samples to better overcome catastrophic forgetting and how to update the memory. Existing replay-based methods usually focus one of them. For the problem of using limited samples, experience replay~(ER) simply replays old saved samples when learning the new task. Some methods improve ER by seeking the most interfered samples in memory to replay~\cite{aljundi2019online} or add regularization terms~\cite{buzzega2020dark}. Some methods design other loss function instead of cross entropy loss~\cite{guo2022online,caccia2022new}. For the problem of updating memory, reservoir sampling~(RS)~\cite{vitter1985random} has been widely adopted by many existing replay-based methods~\cite{buzzega2020dark,gupta2020maml} for memory updating. This memory updating mechanism can guarantee that each sample in the data stream has the same probability of being kept in memory, and subsequently classes with more samples in the data stream will keep more samples in memory. As a result, RS performs poorly in a more realistic environment where the class distribution is highly imbalanced. Gradient-based sample selection~(GSS)~\cite{aljundi2019gradient} selects~(saves) samples to maximize the gradient direction in memory and shows better performance than RS in the imbalanced data stream setting. Class-balancing reservoir sampling~(CBRS)~\cite{chrysakis2020online} proposes class balancing mechanisms for single-label setting to keep an equal number of samples for each class. In this way, the performance of those minor classes can be kept. There are also some works~\cite{DBLP:journals/corr/abs-2204-04763,yoon2021online,tiwari2022gcr,shim2021online} considering how to keep important samples in memory but most of them requires task identities and boundaries. Furthermore, all these methods consider only single-label setting.

When dealing with the online continual learning, augmented graph convolutional netowrk~(AGCN)~\cite{du2022agcn}, class-balanced exemplar selection~(CBES)~\cite{yan2021framework} and partition reservoir sampling~(PRS)~\cite{kim2020imbalanced} take into account the multi-label problem. AGCN considers the relationship between labels and uses a graph network to learn them. However, this method requires task identities and task boundaries to update the graph network. Therefore, it can not be applied to the setting we consider in this work where task boundaries and identity are unavailable during the training process. CBES and PRS consider the same setting as this work. CBES generalizes CBRS to the multi-label setting. However, as demonstrated by the experimental results in this work, this method fails to control the class distribution in memory effectively. PRS has two processes called sample-in/sample-out when a new batch comes and memory is full. In the sample in process, memory decides whether let the current new sample come into the memory or not. If the answer is true, the sample out mechanism will be performed to decide which samples will be moved out. Through this process, PRS can balance the class distribution in memory. However, PRS suffers from slow memory updating speed due to its complicated process. But the speed is important since the agent needs to react quickly to the changes in the real environment.

\section{Methodology}
In this section, we first 
formalize the online continual learning problem with multi-label samples, and then propose our replay-based method OCDM.
\subsection{Problem Definition}
We follow the existing work~\cite{caccia2022new} to formalize the online continual learning setting. Specifically, there is a non-stationary data stream $\mathcal{S}$ containing a sequence of datasets for different tasks. At time $t$, the agent receives a mini-batch of samples $\mathcal{B}_{t}=\{(\bm{x}_{i}^{t},\bm{y}_{i}^{t})\}_{i=1}^{b_{t}}$ from the data stream, where $\bm{x}_{i}^{t}$ is an input sample and $\bm{y}_{i}^{t}$ is a multi-hot label vector. $b_{t}$ is the batch size at time $t$ and it is usually small. Each sample in $\mathcal{B}_{t}$ is drawn from a distribution $D_{s}$. $D_{s}$ is not static and may suddenly change to $D_{s+1}$ when $t$ increases. When the change happens, some new classes will be presented to the agent and some old classes will disappear, which may cause catastrophic forgetting. Furthermore, the task identities are not revealed to the agent~(no task identities), so the agent does not know when the old task ends and when the new task begins~(no task boundaries).  

The goal in online continual learning is to make the agent $f(\bm{x},\theta)$ guarantee its performance on the old distribution while learning on the new distribution. Specifically, at time step $T$, the agent aims to minimize the loss
\begin{eqnarray}
  \mathbb{E}_{s\in [S_{T}]}\left[\mathbb{E}_{(\bm{x},\bm{y})\sim D_{s}}[l(f(\bm{x};\theta),\bm{y})]\right],
\end{eqnarray}
where $[S_{T}]=\{1,2,...,S_{T}\}$ denotes all the tasks the agent has visited until time $T$.

For the replay-based method, there is a memory $\mathcal{M}$, which is used to keep a small part of samples from the data stream. Following the existing online continual learning methods~\cite{DBLP:conf/nips/JinSDR21,kim2020imbalanced}, we assume the memory size is fixed as $M$. At each time step $t$, the agent receives the new observation $\mathcal{B}_{t}$ and should complete two processes. The first process is updating the parameters $\theta$ with $\mathcal{B}_{t}$ and the old data sampled from $\mathcal{M}$. The second process is using the new observation $\mathcal{B}_{t}$ to update $\mathcal{M}$. We primarily focus on the multi-label setting, and we have mentioned in Section~\ref{sec:introduction} that updating memory in multi-label setting will present challenges that single-label setting does not have. Therefore, like some of the existing methods~\cite{kim2020imbalanced,aljundi2019gradient,chrysakis2020online}, memory updating is our primary concern.

\subsection{OCDM}
Like the existing memory updating methods with fixed memory size~\cite{vitter1985random,aljundi2019gradient,kim2020imbalanced}, OCDM consists of two stages. In the first stage, the memory is not full. All the observed mini-batches from the data stream will be added into memory directly in order to make full use of memory space. When the memory is full, OCDM moves to the second stage. In this stage, the agent needs to update the memory according to the new observations and the distribution in memory to keep important information.

When the $t$-th observation $\mathcal{B}_{t}$ with size $b_{t}$ comes and the memory is 
full, the process of memory updating can be regarded as selecting $M$ samples 
from $M+b_{t}$ samples to be kept in $\mathcal{M}$, and then removing the remaining $b_{t}$ 
samples. Different from PRS that treats the samples kept in $\mathcal{M}$ and new 
samples in $\mathcal{B}_{t}$ differently, OCDM treats all these samples equally. Specifically, OCDM first sets the target distribution $\bm{p}$, which we want to control the class distribution in memory to be. Here, $\bm{p}$ is denoted as
\begin{eqnarray}
  & \bm{p}=[p_{1},p_{2},...,p_{C-1},p_{C}],  \nonumber \\
  & p_{i}\geq 0,\hspace{0.2cm} \sum_{i=1}^{C}p_{i}=1, 
\end{eqnarray}
where $C$ represents the total number of classes kept in $\mathcal{M}\cup \mathcal{B}_{t}$. Then OCDM formalizes the process of
selecting $M$ samples from $M+b_{t}$ samples as the following optimization problem:
\begin{eqnarray}\label{eq1}
        \min_{\Omega}~~&Dist(\bm{p}_{\Omega}, \bm{p})\nonumber\\
        s.t.~~~&\Omega\subseteq  [M+b_{t}],\\
        &|\Omega|=M,\nonumber
\end{eqnarray}
where $Dist(\cdot,\cdot)$ represents a function that measures the distance 
between two distributions, such as KL-divergence. $[M+b_{t}]=\{1,2,...,M+b_{t}\}$ represents the indices 
of $\mathcal{M}\cup\mathcal{B}_{t}$, $\Omega$ denotes the indices of the selected dataset.
$\bm{p}_{\Omega}=[p_{\Omega,1}, p_{\Omega,2},...,p_{\Omega,C}]$ represents the class distribution of the selected dataset and its $j$-th component can be represented as
\begin{eqnarray}\label{eq1.5}
        p_{\Omega,j}=\frac{m_{\Omega,j}}{\sum_{i}m_{\Omega,i}},
\end{eqnarray}
where $m_{\Omega,j}$ is the number of samples belonging to the $j$-th 
class in the selected dataset.

\begin{figure*}[t] 
  \vskip 0.0in
  \begin{center}
  \centerline{\includegraphics[width=1\linewidth]{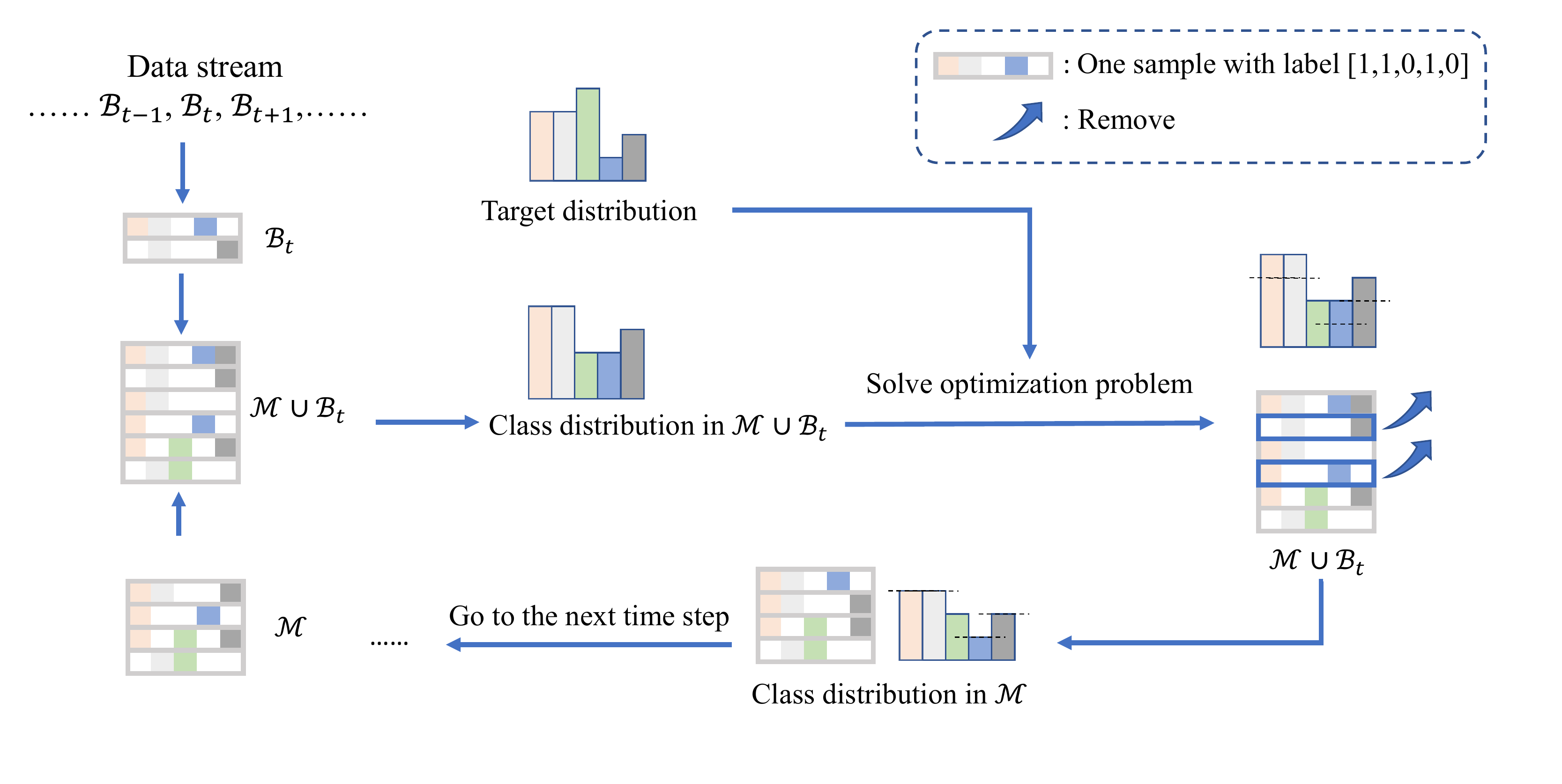}}
  \caption{The process of OCDM. At each step the memory receives a batch of 
  new samples and then removes some samples to minimize the distance between the current 
  class distribution and the target class distribution.}
  \label{ocdm-process}
  \end{center}
\end{figure*}

Solving the optimization problem in~(\ref{eq1}) means 
selecting a subset $\{(\bm{x}_{i},\bm{y}_{i})|~i\in \Omega,(\bm{x}_{i},\bm{y}_{i})\in \mathcal{M}\cup \mathcal{B}_{t}\}$ 
with $M$ elements such that the class distribution $\bm{p}_{\Omega}$ of this selected dataset is closest to the target distribution $\bm{p}$ among all the possibilities. We show the whole process in Figure~\ref{ocdm-process}. In the following two subsections, we will introduce the target distribution $\bm{p}$ and the selection strategy based on the problem in~(\ref{eq1}).

\subsubsection{Target Distribution}
We adopt similar techniques in PRS to formulate our target distribution. That is, the agent uses the running class 
frequency to set the target class distribution in $\mathcal{M}$. Specifically, the agent counts the running frequency 
of each class in the data stream during the training. Here, let $n_{i}$ represent the running frequency for the $i$-th class. When 
the $t$-th batch of new samples comes, the agent sets the target distribution for the $i$-th class in memory as follows:
\begin{eqnarray}\label{eq2}
        p_{i}=\frac{(n_{i})^{\rho}}{\sum_{i=1}^{C}(n_{j})^{\rho}},
\end{eqnarray}
where $\rho$ is a power of allocation. Please note that 
when $\rho=1$, the number of samples kept in memory for each class is 
proportional to the number of samples in the data stream. Therefore, if class $c_{1}$ 
consists of much more samples than class $c_{2}$ in the data stream, setting $\rho=1$ will make 
the memory keep much more samples with class $c_{1}$ than class $c_{2}$. When $\rho=0$, the 
target distribution puts an equal allocation for all the classes, so the memory is updated to 
have the same number of samples in memory for all the classes. 

For the single-label setting without any prior knowledge, keeping an equal number of samples in memory for each class is an 
ideal choice and has been adopted by many existing relay-based methods~\cite{de2021continual,lopez2017gradient,prabhu2020gdumb}. Under the online continual learning setting with an imbalanced data stream, existing works~\cite{chrysakis2020online,DBLP:journals/corr/abs-2204-04763} have shown that keeping class distribution in memory balanced becomes more critical to keep the agent's performance on those old classes with too few samples in the data stream. Existing work PRS has shown that balanced class distribution in 
memory is also ideal for the agent to get desired performance in the multi-label setting. 

In addition, balancing the class distribution in memory is also consistent with the researches in long-tail problem, including single-label long-tail problem~\cite{DBLP:conf/iclr/KangXRYGFK20} and multi-label long-tail problem~\cite{DBLP:conf/eccv/WuH0WL20}. Specifically, the data samples saved in the memory participate in the update of the model, so the memory is part of the training set. Therefore, balancing the class distribution in memory also ensures the class distribution in training set balanced to a certain extent.
Based on the above discussion, we will set $\rho$ to 0 by default unless otherwise stated. In the experiment, we will 
change $\rho$ between 0 and 1 to show the effect of $\rho$ on the agent's performance.

\subsubsection{Selection Strategy}\label{selection strategy}

One challenge we have not tackled is how to solve the optimization problem in~(\ref{eq1}). In some cases of single-label settings, 
solving this problem is trivial. For example, suppose $\rho$ is 0 and memory size is $M=1000$, and there are three 
classes $c_{1}$, $c_{2}$ and $c_{3}$ in $\mathcal{M}\cup\mathcal{B}_{t}$. $c_{1}$ has 300 samples, $c_{2}$ has 500 samples and $c_{3}$ has 300 samples. Obviously, 
we can get the optimal solution in~(\ref{eq1}) by randomly removing 100 samples from $c_{2}$. However, 
this operation is impossible for the multi-label setting because some samples with class label $c_{2}$ may also have class label $c_{1}$ or class label $c_{3}$.
Therefore, changing the number of samples belonging to $c_{2}$ may influence the number of samples belonging to $c_{1}$ or $c_{3}$.

To solve~(\ref{eq1}) in the multi-label setting, one of the straightforward ways
is to traverse all the possibilities and find the optimal value. 
However, the total number of possibilities in~(\ref{eq1}) is $\binom{M+b_{t}}{M}$. Taking our experimental setting as an example, the size of the memory $M$ is set to 1000, and the batch size of the 
data stream $b_{t}$ is set to 10. In this case, we can get:
\begin{eqnarray}\label{eq3}
      \binom{M+b_{t}}{M}&&=\binom{M+b_{t}}{b_{t}}\geq \left(\frac{M+b_{t}}{b_{t}}\right)^{b_{t}}\nonumber\\
      &&=101^{10}\geq 10^{20}.
\end{eqnarray}
This shows that obtaining the optimal solution through traversing
methods is impossible.

In order to tackle with multi-label setting, OCDM uses a greedy algorithm. Specifically, 
the purpose of~(\ref{eq1}) is to select $M$ samples from $M+b_{t}$ samples. Equivalently, the purpose 
of~(\ref{eq1}) can also be restated as deleting $b_{t}$ samples from $M+b_{t}$ samples. Through the above 
analysis, we have known that it is too complicated to delete $b_{t}$ samples from $M+b_{t}$ samples at 
one time. Hence, we divide the deletion process into $b_{t}$ subprocesses, and delete one 
sample in each subprocess. Assuming $\Omega_{0}=[M+b_{t}]$ is the indices of the whole dataset $\mathcal{M}\cup \mathcal{B}_{t}$ 
available to the agent, and in the $k$-th~($1\leq k\leq b_{t}$)
subprocess, $\Omega_{k}$ is selected from $\Omega_{k-1}$ by removing one sample $(\bm{x}_{i_{k}},\bm{y}_{i_{k}})$,
where $\Omega_{k}=\Omega_{k-1}\backslash \{i_{k}\}$. The deleted index $i_{k}$ needs to satisfy
\begin{eqnarray}\label{eq4}
      i_{k}=\mathop{\arg\min}_{j\in \Omega_{k-1}}Dist(\bm{p}_{\Omega_{k-1}\backslash \{j\}},\bm{p}),
\end{eqnarray}
where $\bm{p}_{\Omega_{k-1}\backslash \{j\}}$ represents the class distribution of the subsets 
with indices $\Omega_{k-1}\backslash\{j\}$. We show the whole process of our method in Algorithm~\ref{alg0}.

\begin{algorithm}[t]
  \caption{Optimizing Class Distribution in Memory~(OCDM)}
  \label{alg0}
\begin{algorithmic}
  \STATE {\bfseries Input:} data stream $\mathcal{S}$, memory $\mathcal{M}$, memory size $M$.
  \WHILE {data stream $\mathcal{S}$ is not over}
  \STATE {Get a batch of new samples $\mathcal{B}_{t}=\{(\bm{x}_{i},\bm{y}_{i})\}_{i=1}^{b_{t}}$ from $\mathcal{S}$;}
  \IF {$\mathcal{M}$ is not full}
  \STATE {Get the remaining space $r$ in $\mathcal{M}$;}
  \STATE {Select $\min(b_{t},r)$ samples $\mathcal{V}_{t}$ from $\mathcal{B}_{t}$ randomly and save in $\mathcal{M}$;}
  \STATE {$\mathcal{B}_{t}\leftarrow \mathcal{B}_{t}\backslash \mathcal{V}_{t}$;}
  \STATE {$b_{t}\leftarrow b_{t}-|\mathcal{V}_{t}|$;}
  \ENDIF
  \IF {$b_{t}>0$}
  \STATE {$\Omega_{0}\leftarrow[M+b_{t}]$;}
  \FOR {$k\in[1,2,...,b_{t}]$}
  \STATE {Get index $i_{k}$ from $\Omega_{k-1}$ according to~(\ref{eq4});}
  \STATE {$\Omega_{k}\leftarrow\Omega_{k-1}\backslash\{i_{k}\}$;}
  \ENDFOR
  \STATE {Select $M$ samples with indices $\Omega_{b_{t}}$ as a new memory and remove the remaining samples;}
  \ENDIF
  \ENDWHILE
\end{algorithmic}
\end{algorithm}



For a sequence of length $n$, the time complexity of finding the sample with the smallest objective value in~(\ref{eq4}) is $O(n)$. Hence, the total time complexity of updating memory $\mathcal{M}$ with the new 
batch $\mathcal{B}_{t}$ is
\begin{eqnarray}\label{eq5}
    &&O(\sum_{i=0}^{b_{t}-1}(M+b_{t}-i))\nonumber\\
    &&=O(b_{t}(M+b_{t})-\frac{b_{t}(b_{t}-1)}{2})\nonumber\\
    &&=O(b_{t}M).
\end{eqnarray} 
The second equation in~(\ref{eq5}) is derived from $M\gg b_{t}$ in common conditions. 
This shows that the time complexity of using greedy algorithm to solve~(\ref{eq1}) is linear.
Note that when all the samples have only one label and $\rho$ is set to 0 in~(\ref{eq2}), our selection strategy can get the optimal 
solution because each deleted sample selected by~(\ref{eq4}) must belong to the class with the maximal number of samples in memory.


\section{Experiments}
We compare our method OCDM with other continual learning methods under the online continual learning setting with multi-label 
samples in the data stream.

\subsection{Experimental Settings}\label{sec4.1}
\subsubsection{Dataset}
Following the existing work PRS~\cite{kim2020imbalanced}, we adopt two widely used multi-label datasets MSCOCO~\cite{lin2014microsoft} and NUS-WIDE~\cite{chua2009nus} to construct multi-label continual learning data streams. We do not directly use the datasets COCOseq and NUS-WIDEseq in PRS~\cite{kim2020imbalanced} as our main experimental datasets because we find that the proportion of samples with only one label in the constructed datasets of PRS is too high. Too many samples with one label make the agent only need to pay attention to single-label samples to control the class distribution in memory, and hence the problem can be transformed into a single-label problem. While a high proportion of samples with multiple labels make the method that pays attention to single-label perform poorly. We will verify this point of view in the experiment.

\begin{figure}[t]
  \centering
  \begin{subfigure}{1\linewidth}
  \centering
  \includegraphics[width=1\linewidth]{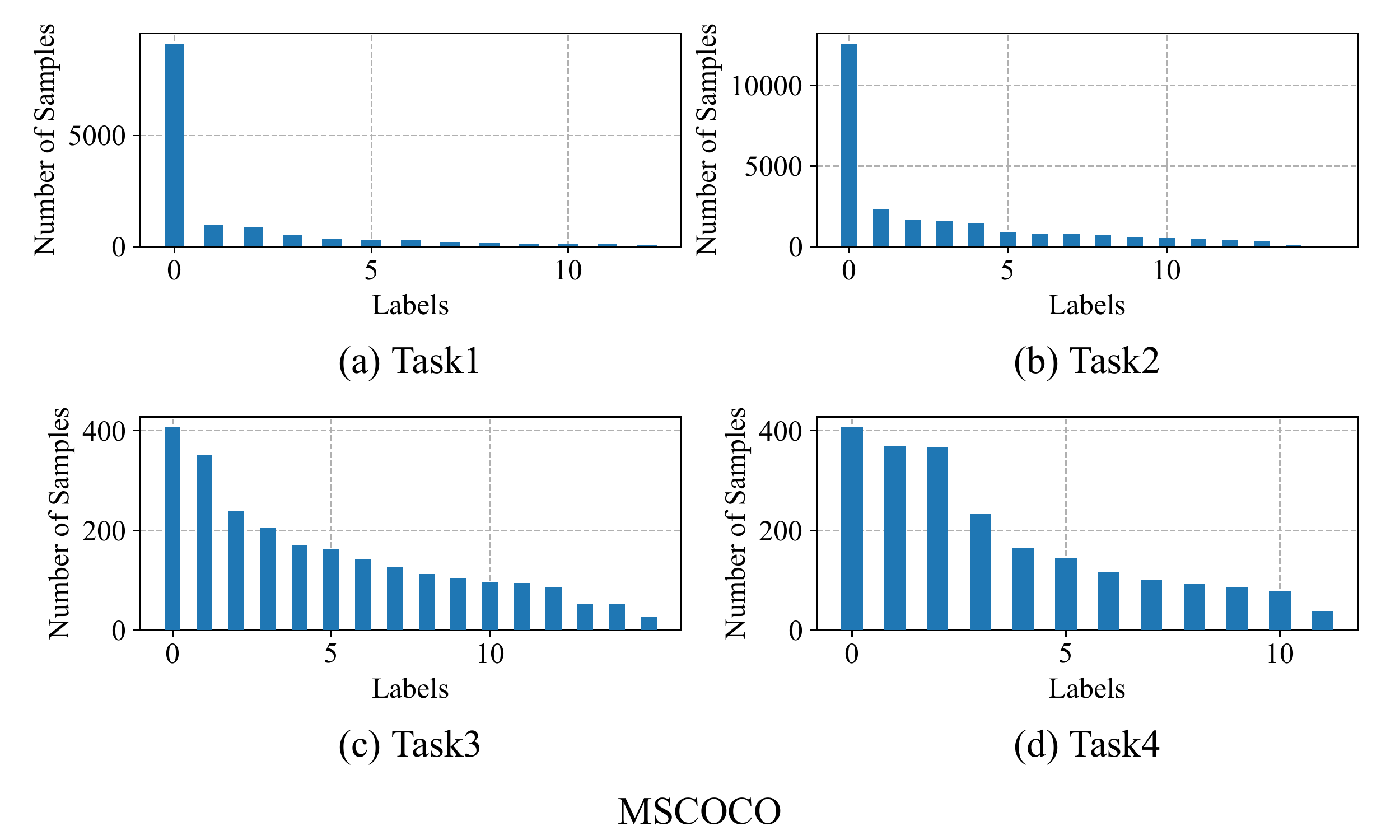}
  \end{subfigure}
  \caption{The class distribution of different tasks in MSCOCO data stream.}
  \label{data stream-mscoco}
\end{figure}

\begin{figure}[t]
  \centering
  \begin{subfigure}{1\linewidth}
  \centering
  \includegraphics[width=1\linewidth]{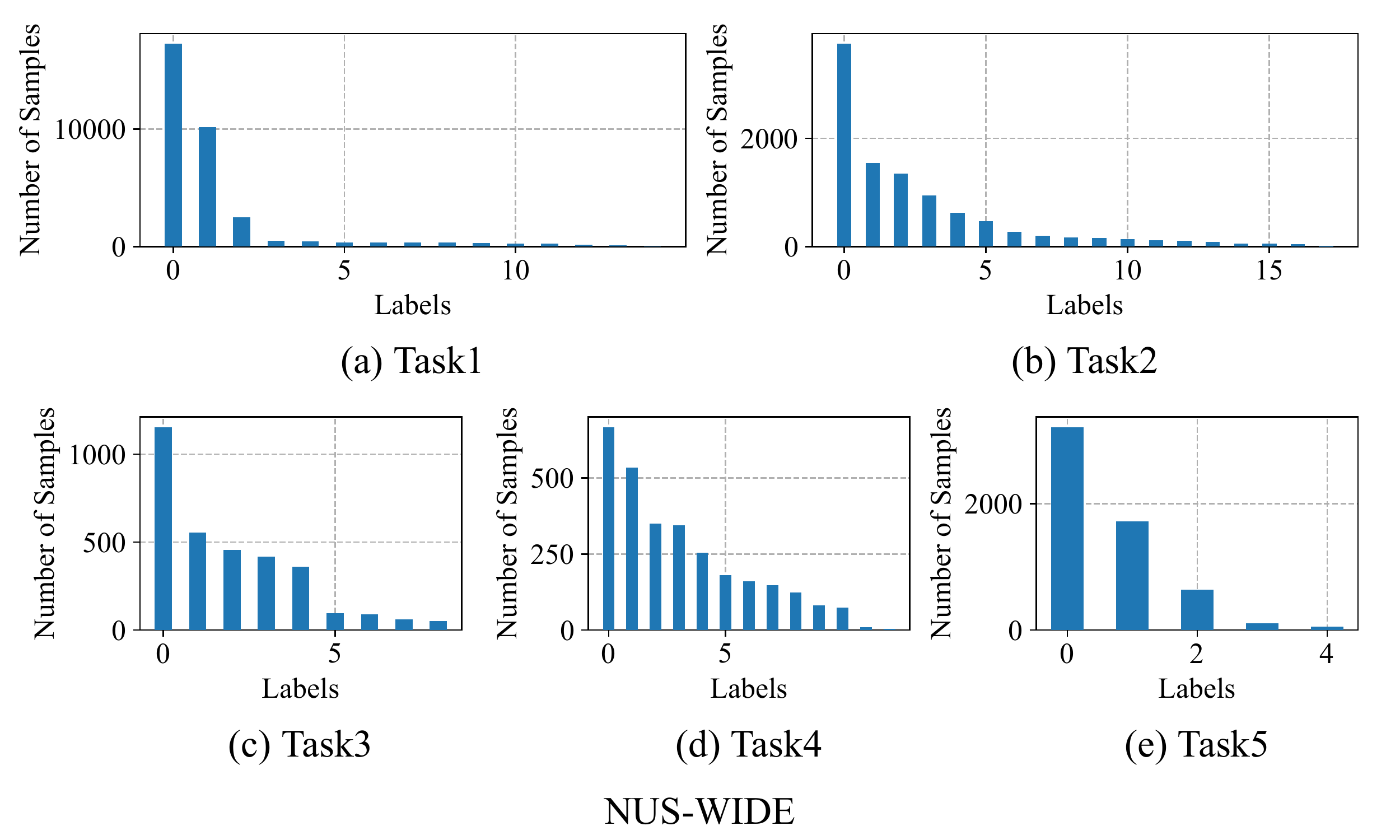}
  \end{subfigure}
  \caption{The class distribution of different tasks in NUS-WIDE data stream.}
  \label{data stream-nuswide}
\end{figure}

To increase the proportion of samples with multiple labels, we create data streams based on MSCOCO and NUS-WIDE. We first use the samples with multiple labels to create the training datasets of several tasks. The method of constructing the training dataset of each task is similar to the existing continual learning methods~\cite{kim2020imbalanced,aljundi2019online}. Specifically, we assign several different classes to each task, and the training dataset of each task only contains the samples belonging to the assigned classes. Then, the remaining samples not included in training datasets are used to construct test sets. Like COCOseq and NUS-WIDEseq, we keep the test datasets balanced. Specifically, the test dataset of each task has 50 samples for each class. During this process, some classes are removed due to the insufficient number of samples for testing. Although removing some classes makes some samples in the training datasets have only one label, the samples with more than one label still account for the majority of the training datasets. Finally, the training datasets of all the tasks are added to a data stream one by one in a specific order. For MSCOCO, we construct four tasks with 44 classes in total and 26617 samples in the data stream. For NUS-WIDE, we construct five tasks with 44 classes in total and 28650 samples in the data stream. In Figure~\ref{data stream-mscoco} and Figure~\ref{data stream-nuswide}, we give the class distribution for these data streams. We can find that the class distributions in these data streams are highly imbalanced.

To show the proportion of multi-label samples in different data streams, we propose Multi-Label Ratio~(MLR). For any class $c$, MLC can measure the proportion of multi-label samples with label $c$. Specifically, the MLR of class $c$ is defined as
\begin{eqnarray}
  {\rm MLR}(c)=\frac{M_{c}}{N_{c}},
\end{eqnarray}
where $N_{c}$ denotes the number of samples with label $c$. $M_{c}$ denotes the number of samples with label $c$ and with multiple labels. We use average-MLR~(AMLR) to denote the average of the MLR of all classes in a dataset. Specifically, AMLR of a dataset $\mathcal{D}$ can be denoted as
\begin{eqnarray}
  {\rm AMLR}(\mathcal{D})&&=\frac{1}{C}\sum_{c\in\mathcal{C}_{\mathcal{D}}}{\rm MLR}(c)\nonumber\\
  &&=\frac{1}{C}\sum_{c\in\mathcal{C}_{\mathcal{D}}}\frac{M_{c}}{N_{c}},
\end{eqnarray}
where $\mathcal{C}_{\mathcal{D}}$ denotes the set of classes in the dataset $\mathcal{D}$. In Table~\ref{task infom1}, we show AMLR for each task in different data streams. We can find that AMLR of the data streams constructed in this paper is much larger than the data streams used in the PRS~\cite{kim2020imbalanced}. This shows that the proportion of samples with multiple labels in the data streams we construct is much higher than the data streams used in PRS.

\begin{table}[t]
  \renewcommand\arraystretch{1}
  \renewcommand\tabcolsep{7.5pt}
  \setlength{\belowcaptionskip}{0.1cm}
  \caption{Average multi-label ratio of each tasks over different data streams.}
  \label{task infom1}
  \centering
  \begin{tabular}{lcccccccc}
    \toprule
    & \multicolumn{2}{c}{MSCOCO}& \multicolumn{2}{c}{NUS-WIDE}\\
    \midrule
                & COCOseq    & Ours & NUS-WIDEseq    & Ours       \\
    \midrule  
    task1       & 58.60$\%$   &	95.53$\%$ & 49.33$\%$   &	 98.08$\%$  \\
    task2       & 55.33$\%$   &	89.12$\%$ & 71.06$\%$   &	 99.15$\%$  \\
    task3       & 15.42$\%$   &	84.33$\%$ & 45.95$\%$   &  88.73$\%$  \\
    task4       & 23.46$\%$   &	64.54$\%$ & 12.44$\%$   &	 95.14$\%$  \\
    task5       &     -       &     -     & 6.19$\%$    &  83.71$\%$  \\
    task6       &     -       &     -     & 52.78$\%$   &      -      \\
    Total       &       42.1$\%$      &      85.6$\%$     &     38.8$\%$        &       96.4$\%$      \\
    \bottomrule
  \end{tabular}
\end{table}

Since the multi-label datasets always have an imbalanced class distribution, we follow the existing work PRS and divide all classes into three parts: majority, moderate, and minority, according to the number of samples in each class. Specifically, classes with a sample size of more than 600 are classified as majority, those with less than 100 are classified as minority, and those with a sample size between 100 and 600 are classified as moderate.

\subsubsection{Baseline and Evaluation Metric} We compare our methods with different baselines including 
RS~\cite{vitter1985random}, PRS~\cite{kim2020imbalanced}, 
GSS-Greedy~\cite{aljundi2019gradient} and CBES~\cite{yan2021framework}. RS, PRS, GSS-Greedy and CBES are the typical or state-of-the-art replay-based continual learning methods that require no task identities or boundaries. Besides, we also add \emph{multitask}, \emph{finetune}, and \emph{onlyone} to compare with our method. 
\emph{Multitask} learns the model with all the old data when a new task is coming, which could be regarded as the upper bound 
of all the continual learning methods. \emph{Finetune} learns the model without any memory, which could be regarded as the lower 
bound of all the methods. 
\emph{Onlyone} is the same as OCDM except that it ignores the samples with multiple labels and only 
uses the samples with one label to update the memory.

We report several commonly used multi-label classification metrics including the average overall F1~(OF1), 
per-class F1~(CF1), and mean average precision~(mAP). 

\subsubsection{Training Details and Architecture}
Following the existing online continual learning methods~\cite{aljundi2019online,aljundi2019gradient}, in all the experiments, 
we set both the data batch size and replay size as 10 for data stream and memory, respectively. We set $\rho=0$ unless otherwise 
stated. Furthermore, we perform data augmentation with random crops and horizontal flips to the samples in the data stream and 
memory for all the baselines and our methods. To keep the comparison fair and follow PRS~\cite{kim2020imbalanced}, we use binary 
cross entropy~(BCE) as the loss function for all the methods, and no regularization is used. 

It is common for recent multi-label methods to fine-tune a pre-trained model to a target dataset. We thus choose ResNet101 
pre-trained on ImageNet as the classifier of the agent to follow PRS~\cite{kim2020imbalanced}. We train the backbones using Adam optimizer with learning 
rate $lr=0.0001$, running averages of gradient and its square $(\beta_{1}=0.9,\beta_{2}=0.999)$, numerical stability 
term $\epsilon=0.0001$ which is used by PRS~\cite{kim2020imbalanced}. We perform all experiments on four NVIDIA TITAN Xp GPUs.

\begin{table*}[t]
  \renewcommand\arraystretch{1}
  \vspace{-0.1cm}
  \renewcommand\tabcolsep{7.4pt}
  \setlength{\belowcaptionskip}{0.1cm}
  \caption{Results of the continual learning methods on MSCOCO with memory size being 1000. }
  \label{tab0}
  \centering
  \begin{tabular}{lccccccccccccc}
    \toprule
                & \multicolumn{3}{c}{Majority} & \multicolumn{3}{c}{Moderate}& \multicolumn{3}{c}{Minority}& \multicolumn{3}{c}{Total}& Time~(s)\\
    \cmidrule(r){2-4}\cmidrule(r){5-7}\cmidrule(r){8-10}\cmidrule(r){11-13}
    Methods     & CF1 & OF1 & mAP & CF1 & OF1 & mAP & CF1 & OF1 & mAP & CF1 & OF1 & mAP & 1000 steps \\
    \midrule  
    Multitask   & 64.1& 45.0& 68.5& 56.7& 53.1& 57.4& 32.7& 30.6& 49.3& 58.7& 48.6& 59.5& - \\
    \midrule 
    Finetune    & 4.1 & 4.5 & 30.6& 12.8& 20.2& 22.8& 8.5 & 14.3& 12.5& 13.6& 16.9& 23.8& - \\
    RS          & 52.1& 38.0& 52.5& 26.2& 25.8& 34.6& 8.3 & 10.7& 12.1& 31.7& 29.0& 36.9& \textbf{656.0} \\
    GSS-Greedy  & 44.8& 36.7& 50.8& 27.6& 26.3& 33.5& 7.1 & 11.9& 13.7& 30.4& 28.0& 36.0& 6040.7 \\
    CBES        & \textbf{54.8}& \textbf{55.5}& \textbf{60.4}& 27.7& 29.1& 35.2& 4.7 & 8.7 & 15.1& 32.6& 34.7& 39.8& 689.0 \\
    PRS         & 48.0& 47.8& 51.0& 39.7& 35.4& 40.3& 8.1 & 8.8 & 21.7& 38.8& 36.5& 41.1& 1215.0 \\
    Onlyone     & 43.6& 47.5& 51.9& 22.1 & 26.4& 29.7& 0.0& 0.0& 12.5& 26.3& 29.6& 33.8& - \\
    OCDM~(ours) & 47.6& 47.3& 51.6& \textbf{40.6} & \textbf{37.8} & \textbf{42.3} & \textbf{21.0} & \textbf{20.9} & \textbf{26.22} & \textbf{40.5} & \textbf{38.9} & \textbf{43.0} & 673.0 \\
    \bottomrule
  \end{tabular}
\end{table*}

\begin{table*}[t]
  \renewcommand\arraystretch{1}
  \vspace{-0.1cm}
  \renewcommand\tabcolsep{7.4pt}
  \setlength{\belowcaptionskip}{0.1cm}
  \caption{Results of the continual learning methods on NUS-WIDE with memory size being 1000.}
  \label{tab1}
  \centering
  \begin{tabular}{lccccccccccccc}
    \toprule
                & \multicolumn{3}{c}{Majority} & \multicolumn{3}{c}{Moderate}& \multicolumn{3}{c}{Minority}& \multicolumn{3}{c}{Total}& Time~(s)\\
    \cmidrule(r){2-4}\cmidrule(r){5-7}\cmidrule(r){8-10}\cmidrule(r){11-13}
    Methods     & CF1 & OF1 & mAP & CF1 & OF1 & mAP & CF1 & OF1 & mAP & CF1 & OF1 & mAP & 1000 steps \\
    \midrule  
    Multitask   & 31.3& 19.2& 21.4& 32.4& 31.8& 29.9& 9.3 & 9.8 & 19.7 & 31.4& 23.2& 25.8&  - \\
    \midrule 
    Finetune    & 2.7 & 5.1 & 8.4 & 2.1 & 6.2 & 8.9 & 0.0 & 0.0 & 6.9 & 2.1 & 5.1 & 8.4 & - \\
    RS          & 22.8& 15.4& 16.2& 11.9& 12.4& 18.2& 1.6 & 1.7 & 9.1 & 17.4& 13.3& 16.2& \textbf{578.1} \\
    GSS-Greedy  & 21.4& 16.4& 16.0& 11.6& 12.4& 18.5& 0.2 & 0.2 & 9.1 & 16.7& 13.8& 16.2& 5808.1 \\
    CBES        & \textbf{26.2}& \textbf{20.3}& \textbf{18.8}& 17.0& 16.5& 21.5& 1.6 & 1.7 & 12.4& 19.9& 17.2& 19.3& 592.7 \\
    PRS         & 22.5& 19.0& 17.0& 23.8& 23.1& 22.3& 16.4& 17.5& 17.7& 23.6& 20.7& 20.0& 1175.2 \\
    Onlyone    & 17.4& 16.3 & 15.4& 4.7 & 8.3 & 12.6& 0.0 & 0.0 & 6.2 & 7.7 & 11.3& 12.4& - \\
    OCDM~(ours) & 20.2& 18.8& 17.0& \textbf{29.2}& \textbf{27.5}& \textbf{25.6}& \textbf{20.2}& \textbf{20.5}& \textbf{20.1}& \textbf{25.8}& \textbf{23.3}& \textbf{22.2}& 580.8 \\
    \bottomrule
  \end{tabular}
\end{table*}

\begin{table*}[h!]
  \renewcommand\arraystretch{1}
  \vspace{-0.1cm}
  \renewcommand\tabcolsep{3.5pt}
  \setlength{\belowcaptionskip}{0.1cm}
  \caption{mAP Results of different continual learning methods over COCOseq and NUS-WIDEseq.}
  \label{tab: previous data stream}
  \centering
  \begin{tabular}{lcccccccccc}
    \toprule
                & \multicolumn{5}{c}{COCOseq}&\multicolumn{5}{c}{NUS-WIDEseq}  \\
    \cmidrule(r){2-6}\cmidrule(r){7-11}
    Methods     & Majority & Moderate & Minority & Total & Time~(1000 steps) & Majority & Moderate & Minority & Total & Time~(1000 steps)  \\
    \midrule  
    Onlyone     & 69.97 & 50.28 & 26.88 & 50.66 & - & \textbf{26.79} & 22.75 & 27.59 & 25.45 & -  \\
    RS          & 70.18 & 46.74 & 22.40 & 47.50 & \textbf{594.9} & 26.59 & 20.74 & 16.79  & 20.24 & \textbf{469.1} \\
    CBES        & 66.73 &	44.46 & 17.37 & 44.75 & 596.2 & 26.39& 22.13& 25.34& 24.38 & 524.4   \\
    PRS         & 68.59 & 50.52 & 31.30 & 51.32 & 1128.6 & 25.54 & 23.10  & 28.16 & 25.74 & 801.8 \\
    OCDM~(ours) & \textbf{70.87}  & \textbf{51.67} & \textbf{31.58} & \textbf{52.59} & 602.0 & 25.55 & \textbf{23.44} & \textbf{28.32} & \textbf{25.96} & 486.6 \\
    \bottomrule
  \end{tabular}
\end{table*}

\subsection{Experimental Results}
We perform the experiments on each data stream three times. Table~\ref{tab0} shows the average performance for different methods over the data stream created based on MSCOCO. Table~\ref{tab1} shows the average performance for different methods over the data stream created based on NUS-WIDE. The memory size is set to 1000 for all the replay-based methods. We can find that RS, GSS-Greedy and CBES get competitive results for the majority classes. However, the results for moderate classes and minority classes are not satisfactory for RS, GSS-Greedy and CBES. As for PRS and our method OCDM, the results are competitive for all classes, so the performance for PRS and our method is much better than RS and GSS-Greedy. Furthermore, OCDM outperforms PRS in most cases in terms of accuracy. \emph{Onlyone} only pays attention to the samples with one label when updating the memory, and we can find it performs poorly. In Section~\ref{final memory}, we will show that \emph{onlyone} makes the class distribution in memory imbalanced. Therefore, it is necessary for replay-based methods to pay attention to multi-label samples in our data stream. In Table~\ref{tab: previous data stream}, we also show the experimental results on COCOseq and NUS-WIDEseq used in PRS~\cite{kim2020imbalanced}. We can find that \emph{onlyone} can get competitive results on the COCOseq and NUS-WIDEseq because these two data streams consist of too many single-label samples. We will further verify this view in Section~\ref{final memory}.


\begin{table*}[h]
  \renewcommand\arraystretch{1}
  \renewcommand\tabcolsep{7pt}
  \setlength{\belowcaptionskip}{0.1cm}
  \caption{Results of the continual learning methods on NUS-WIDE and MSCOCO with memory size varing.}
  \label{tab2}
  \centering
  \begin{tabular}{lcccccccccccc}
    \toprule
    & \multicolumn{6}{c}{MSCOCO}& \multicolumn{6}{c}{NUS-WIDE}\\
    \midrule
                & \multicolumn{3}{c}{$M=300$} & \multicolumn{3}{c}{$M=500$}& \multicolumn{3}{c}{$M=300$}& \multicolumn{3}{c}{$M=500$}\\
    \cmidrule(r){2-4}\cmidrule(r){5-7}\cmidrule(r){8-10}\cmidrule(r){11-13}
    Methods     & CF1 & OF1 & mAP & CF1 & OF1 & mAP & CF1 & OF1 & mAP & CF1 & OF1 & mAP  \\
    \midrule  
    RS          & 24.8& 24.6& 30.9& 28.7& 27.0& 35.1& 11.4& 8.3 & 13.3& 12.5& 10.2& 13.6 \\
    GSS-Greedy  & 24.2& 23.7& 29.8& 27.2& 25.7& 31.5& 14.2& 10.2& 14.9& 15.6& 10.7& 15.6 \\
    CBES        & 29.4& 30.6& 34.0& 31.8& 33.8& 36.6& 13.2& 8.7 & 14.4& 13.7& 11.7& 15.9 \\
    PRS         & \textbf{32.3}& \textbf{30.2}& 35.2& 35.4& 33.9& 38.9& 16.3& 12.6& 15.6& 20.4& 17.3& 18.2 \\
    OCDM~(ours) & 31.3& \textbf{30.2}& \textbf{35.8}& \textbf{35.9}& \textbf{35.2}& \textbf{39.5}& \textbf{17.2}& \textbf{13.4}& \textbf{16.2}& \textbf{21.6}& \textbf{17.9}& \textbf{18.9} \\
    \bottomrule
  \end{tabular}
  \vspace{-0.2cm}
\end{table*}

\begin{figure*}[h]
  \centering
  \begin{subfigure}{1\linewidth}
  \centering
  \includegraphics[width=1\linewidth]{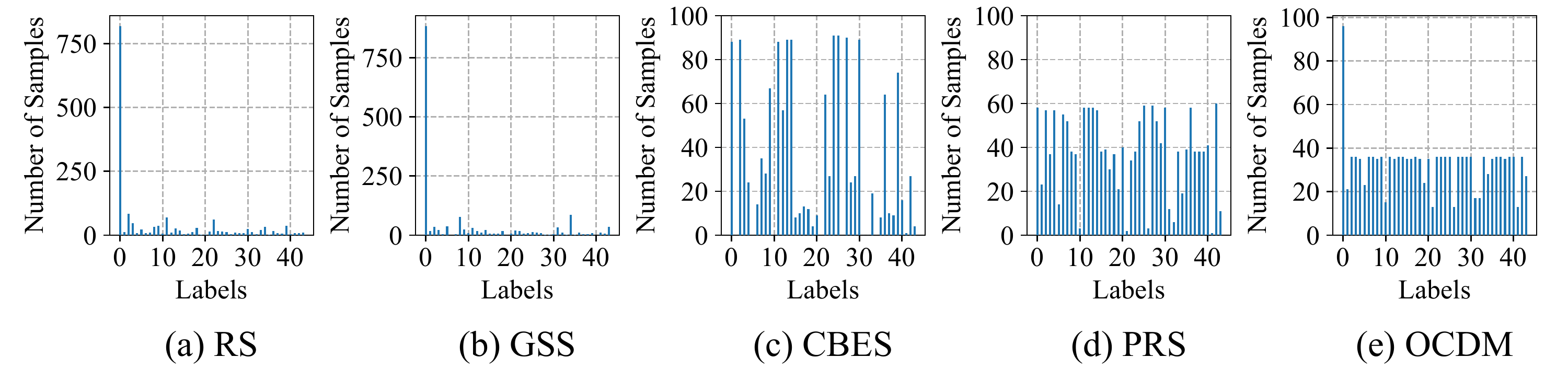}
  \centering
  \includegraphics[width=1\linewidth]{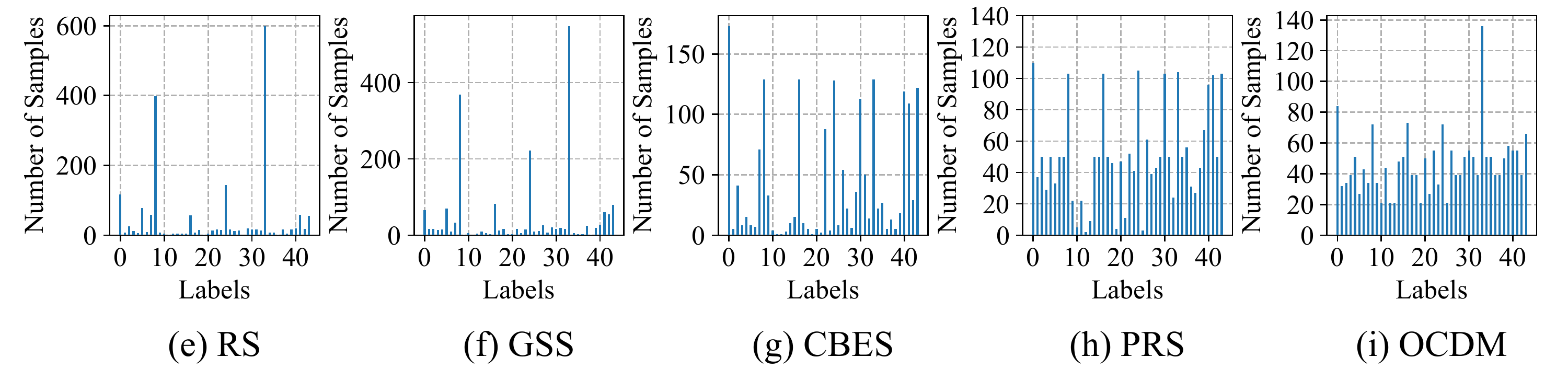}

  \end{subfigure}
  \caption{(a),~(b),~(c) and~(d) show the class distribution in memory after training on MSCOCO with memory size $M=1000$.~(e),~(f),~(g) and~(h) show the class distribution in memory after training on NUS-WIDE with memory size $M=1000$.}
  \label{memory distribution-coco}
\end{figure*}

\begin{figure}[h]
  \centering
  \begin{subfigure}{1\linewidth}
  \centering
  \includegraphics[width=1\linewidth]{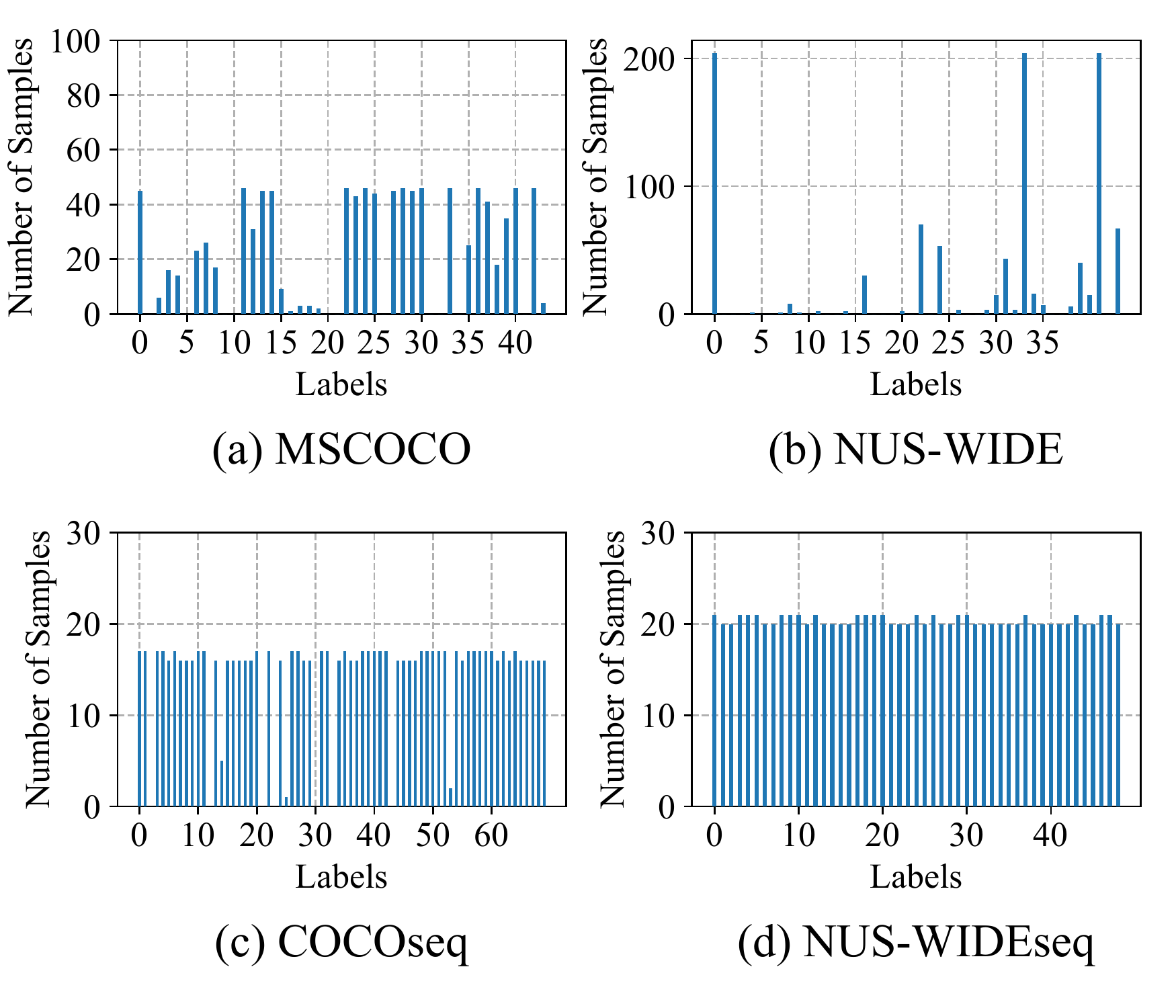}

  \end{subfigure}
  \caption{(a) and~(b) show the class distribution in memory after training on MSCOCO and NUS-WIDE with memory size $M=1000$ and \emph{onlyone} strategy. (c) and~(d) show the class distribution in memory after training on COCO-seq and NUS-WIDE-seq with memory size $M=1000$ and \emph{onlyone} strategy.}
  \label{memory distribution-onlyone}
\end{figure}


Table~\ref{tab0} and Table~\ref{tab1} also report the average time consumed by each replay-based method per 1000 steps in the training process. One step means the agent observes one batch of samples and then updates parameters and memory with these samples. Among them, GSS-Greedy is the slowest one, which is also consistent with the experimental results in existing work~\cite{chaudhry2020continual}. Because the memory update operation of RS is simple and straightforward, RS is the fastest one among all the replay-based methods. Our method OCDM and PRS have the mechanism to control the class distribution in memory, so more time is consumed than RS. However, the speed of OCDM is comparable with that of RS, but PRS is much slower than RS and OCDM due to its complicated process.

\subsubsection{Changing the Memory Size} 
We also follow the existing works~\cite{aljundi2019online,kim2020imbalanced} and vary the memory size in order to show whether the conclusion is consistent in other memory sizes. Specifically, we set the memory size to 300 and 500, and show the results in Table~\ref{tab2}. We can find that the performance of all the methods is lower than that with $M=1000$, which is consistent with intuition. However, our method OCDM still achieves the best performance in most cases. 


\subsubsection{Class Distribution in Memory after Training}\label{final memory}
We show the class distribution in memory for RS, CBES, PRS, and OCDM in Figure~\ref{memory distribution-coco} after training on MSCOCO and NUS-WIDE data streams. We can find the class distribution for RS and GSS-Greedy is highly imbalanced. Specifically, in the MSCOCO data stream, both RS and GSS make 1 class have more than 500 in memory. In the NUS-WIDE data stream, these two methods make 2 classes have more than 400 samples in memory. CBES keeps the distribution of some classes balanced in memory but fails to keep class distribution balanced for others. For PRS and OCDM, the class distribution in memory is relatively balanced, while our method OCDM is more balanced than PRS. Specifically, in the MSCOCO data stream, PRS makes 18 classes have more than 40 samples and 9 classes have less than 20 samples in memory. Our method OCDM makes only 1 class have more than 40 samples and only 6 classes have less than 20 samples in memory. In the NUS-WIDE data stream, PRS makes 11 classes have more than 60 samples, and 6 classes have less than 20 samples in memory. Our method OCDM makes only 6 classes have more than 60 samples and no classes have less than 20 samples in memory.

Furthermore, we also give the class distribution in memory for \emph{onlyone} in Figure~\ref{memory distribution-onlyone}~(a) and Figure~\ref{memory distribution-onlyone}~(b). We can find that \emph{onlyone} also makes the class distribution in memory imbalanced like RS and GSS, which means only focusing on the single-label samples to update the memory fails to control the class distribution in memory. In Figure~\ref{memory distribution-onlyone}~(c) and Figure~\ref{memory distribution-onlyone}~(c), we show the class distribution in memory after training on COCOseq and NUS-WIDEseq with \emph{onlyone}, we can find that \emph{onlyone} can control class distribution in memory on these two data streams. This is why \emph{onlyone} can get competitive performance on COCOseq and NUS-WIDEseq.

\subsubsection{Performance Change during Training}
\begin{figure*}[t]
  \centering
  \begin{subfigure}{1\linewidth}
    \includegraphics[width=1.0\linewidth]{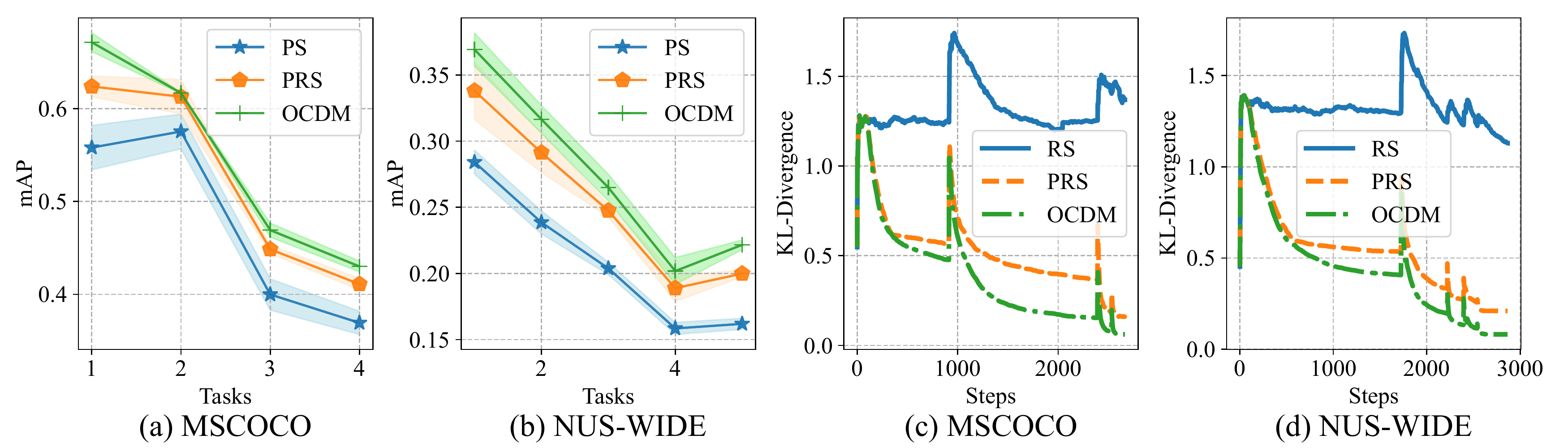}
  \end{subfigure}
  \caption{(a) and~(b) show the variation of mAP over different data streams.~(c) and~(d) show the variation of distance between class distribution in memory and target distribution over different data streams.}
  \label{performance_change}
\end{figure*}

We show the changes in mAP after each task ends in Figure~\ref{performance_change}~(a) and  Figure~\ref{performance_change}~(b). It can be seen that the mAP of all methods gradually decreases as the number of tasks increases in both these two data streams, but our method is consistently better than PRS and RS. This result is important because the data stream may not be terminated in an actual environment. Hence, it is more important to maintain good performance over the whole data stream.

\subsubsection{Distance Change during Training}
We show the distance change between class distribution in memory and target distribution during the whole training on different data streams in Figure~\ref{performance_change}~(c) and Figure~\ref{performance_change}~(d), where KL-divergence is used as the distance function. There are several sharp points in the figure. At these points, the class distribution of the data stream has changed, that is, new classes of samples have appeared. In RS, the distance does not decrease with the increase of the number of steps. On the contrary, the distance in RS gradually increases sometimes. Using PRS and our method OCDM can make the distance decrease gradually, which means the class distribution in the memory gradually converges to the target distribution. Furthermore, OCDM can make the class distribution in memory closer to the target distribution than PRS. These results show that OCDM can control the class distribution in memory better than other methods.

\begin{figure*}[t]
  \centering
  \begin{subfigure}{1\linewidth}
    \includegraphics[width=1.0\linewidth]{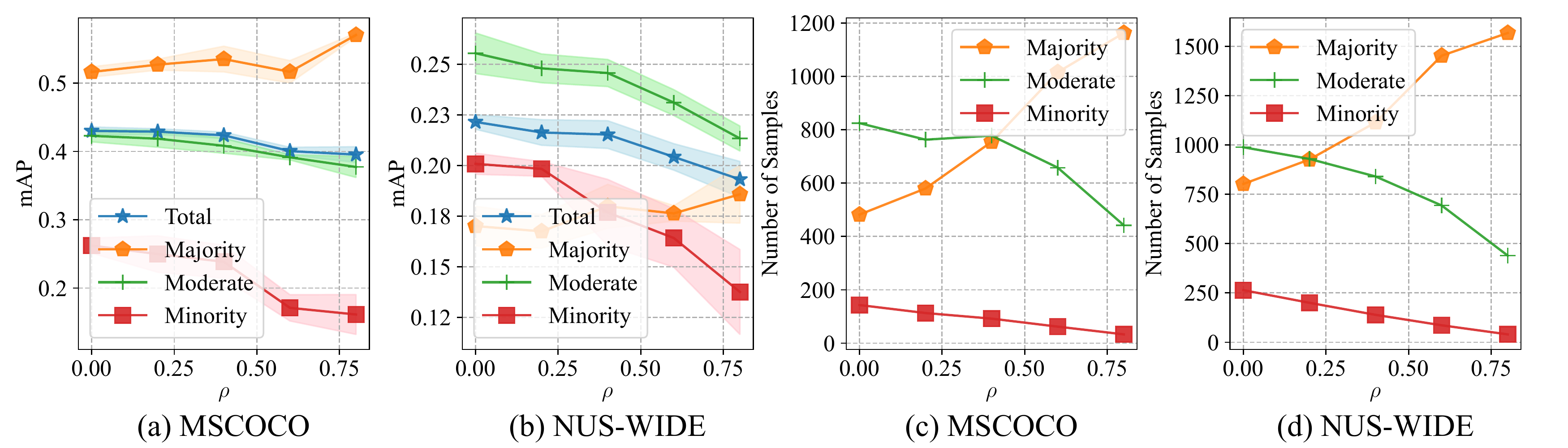}
  \end{subfigure}
  \caption{ (a) and~(b) show the change of OCDM's performance 
  on MSCOCO and NUS-WIDE with difference values of $\rho$; (c) and~(d) show the samples kept in memory~(Using OCDM) after training on MSCOCO and NUS-WIDE with 
  different values of $\rho$.}
  \label{fig3.5}
\end{figure*}
\subsubsection{Changing the Value of $\rho$}
We change the parameter $\rho$ of the target distribution in~(\ref{eq2}) to get different target class distributions in memory and then study the influence on the model's performance. We set $\rho$ to the five values of 0.0, 0.2, 0.4, 0.6, and 0.8. Figure~\ref{fig3.5}~(a) and~(b) show the mAP of OCDM on MSCOCO and NUS-WIDE, respectively. Figure~\ref{fig3.5}~(c) and Figure~\ref{fig3.5}~(d) show the number of samples in memory with different $\rho$.

From Figure~\ref{fig3.5}~(a) and Figure~\ref{fig3.5}~(b), We can find that the model's performance on moderate and minority classes shows decreasing trends with the increases of $\rho$. From Figure~\ref{fig3.5}~~(c) and Figure~\ref{fig3.5}~(d), we can find that the number of samples belonging to moderate and minority classes decreases with the increase of $\rho$. Therefore, the model's performance on moderate and minority classes decreases due to too few samples kept in memory. For the majority classes, we find from Figure~\ref{fig3.5}~(c) and Figure~\ref{fig3.5}~(d) that the model's performance on the majority classes shows increasing trends with the increase of $\rho$. It can be seen from Figure~\ref{fig3.5}~(c) and Figure~\ref{fig3.5}~(d) that the number of samples of the majority classes gradually increases with the increase of $\rho$.

\begin{table}[t]
  \scriptsize
  \renewcommand\arraystretch{1.5}
  \renewcommand\tabcolsep{7pt}
  \setlength{\belowcaptionskip}{0.1cm}
  \caption{Results of the continual learning methods on MSCOCO and NUS-WIDE with different update strategy.}
  \label{tab3}
  \centering
  \begin{tabular}{lcccc}
    \toprule
    & \multicolumn{3}{c}{MSCOCO} & Time~(s) \\
    \cmidrule(r){2-4}
    Methods     & CF1 & OF1 & mAP & 1000 steps \\
    \midrule  
    max    & 33.4 $\pm$ 0.8 & 36.1 $\pm$ 0.6 & 39.6 $\pm$ 1.6 & 663.1  \\
    random & 28.7 $\pm$ 2.0 & 28.7 $\pm$ 1.7 & 33.4 $\pm$ 2.1 & \textbf{659.8}  \\
    OCDM & \textbf{40.5 $\pm$ 1.7} & \textbf{38.9 $\pm$ 1.8} & \textbf{43.0 $\pm$ 0.7} & 674.2 \\
    \bottomrule

    & \multicolumn{3}{c}{NUS-WIDE} & Time~(s) \\
    \cmidrule(r){2-4}
    Methods     & CF1 & OF1 & mAP & 1000 steps \\
    \midrule  
    max    & 19.5 $\pm$ 1.8& 19.5 $\pm$ 0.6& 19.2 $\pm$ 0.7 & 583.9 \\
    random & 8.3 $\pm$ 1.7 & 9.4 $\pm$ 1.0 & 13.2 $\pm$ 0.8 & \textbf{579.6} \\
    OCDM & \textbf{25.8 $\pm$ 0.5}& \textbf{23.3 $\pm$ 0.6}& \textbf{22.2 $\pm$ 0.4} & 582.1 \\
    \bottomrule

  \end{tabular}
\end{table}

\begin{figure}[t]
  \centering
  \begin{subfigure}{1\linewidth}
    \includegraphics[width=1.0\linewidth]{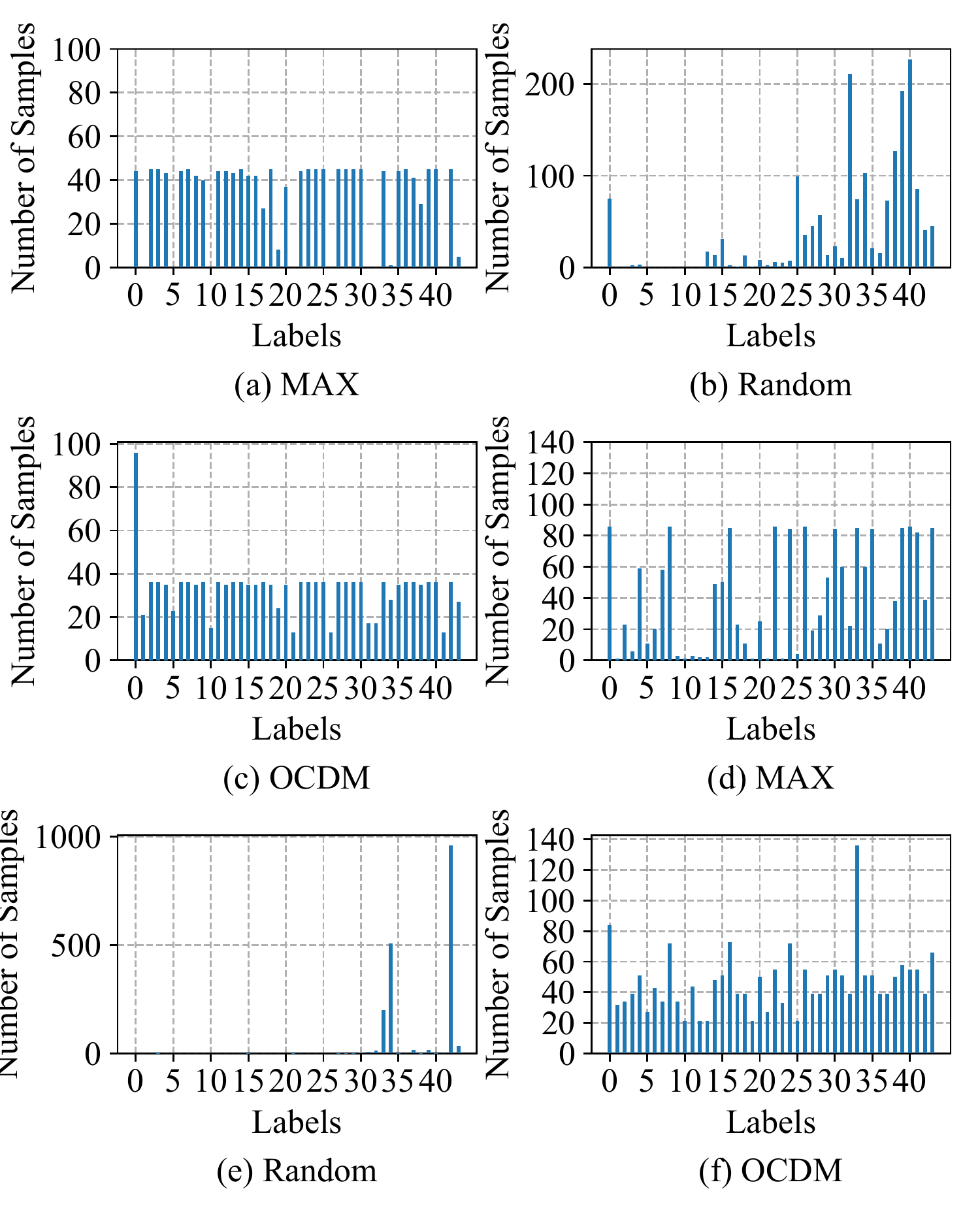}
  \end{subfigure}
  \caption{(a),~(b) and~(c) show the class distribution in memory after training on MSCOCO with different strategies.
    (d),~(e) and (f) show the class distribution in memory after training on NUS-WIDE with different strategies.}
  \label{fig:different strategies}
\end{figure}
\subsubsection{Verifying Effectiveness of Update Strategy}
In order to further verify the effectiveness of our greedy strategy, we compare two strategies that do not use the greedy algorithm to delete samples when updating memory, including random and max. Random strategy randomly selects a sample to delete each time in the process of updating the memory. Max strategy randomly selects a sample belonging to the class with the maximal number of samples in $\mathcal{M}\cup\mathcal{B}_{t}$ to delete each time in the process of updating the memory. The experimental results on MSCOCO and NUS-WIDE with memory $M=1000$ are in Table~\ref{tab3}. We can see that these two methods get similar time consumption as our OCDM for updating the memory. However, they get much lower accuracy than our method. In Figure~\ref{fig:different strategies}, we also show the class distribution in memory over different methods after training on the MSCOCO data stream and NUS-WIDE data stream. We can find that random and max make the class distribution in memory more imbalanced than our OCDM.

Random strategy has the worst result because it cannot control the class distribution in memory. Specifically, when the model learns on a new task, random strategy will remove $b_{t}$ samples from $\mathcal{M}\cup \mathcal{B}_{t}$ for each new batch of samples. With this strategy, the ratio between the number of old samples in memory after memory updating and the number of old samples in memory before memory updating is $\frac{M}{M+b_{t}}$, that is $\frac{100}{101}$ in this experiment. Repeating this process $k$ times will make this ratio become $(\frac{100}{101})^{k}$. With $k$ increasing, the number of samples belonging to old tasks will reduce to 0 quickly, so this strategy fails to keep performance on old tasks. In Figure~\ref{fig:different strategies}~(b) and~(e), we verify this claim. Specifically, the classes in Figure~\ref{fig:different strategies}~(b) and~(e) have been arranged in the order in which they appear in the data stream. The classes on the left belong to the old tasks, and the classes on the right belong to the new tasks. We can see that there are far more new~(right) classes stored in memory than old~(left) classes.

As for max strategy, removing a sample belonging to the labels with the maximal number of samples in $\mathcal{M}_{t}\cup\mathcal{B}_{t}$ is equivalent to our greedy strategy in the single-label setting. However, in the multi-label setting, this strategy may influence the labels which do not contain the maximal number of samples. Therefore, max strategy fails to control the class distribution in memory and performs worse than OCDM. Consider an extreme setting where there are two classes $c_{1}$ and $c_{2}$. $c_{1}$ has 10 samples and $c_{2}$ has 5000 samples. Each sample belonging to $c_{1}$ also belongs to $c_{2}$. Then, using max strategy will make the ratio between the number of samples belonging to $c_{1}$ in memory and the number of samples belonging to $c_{2}$ in memory always be 1:500. In Figure~\ref{fig:different strategies}~(a) and~(d), we can find that max can ensure balance for some classes. However, for some classes, the number of samples saved by max is much less than other classes, because these classes are dominated by other classes in the data stream. The same phenomenon happens with CBES. Specifically, CBES also has the process of randomly selecting a sample belonging to the class with the maximal number of samples in memory to delete. Therefore, CBES also cannot effectively control the distribution of memory when some classes are dominated by other classes.

\subsubsection{Randomizing Task Order}
\begin{figure}[h]
  \centering
  \begin{subfigure}{1\linewidth}
    \includegraphics[width=1.0\linewidth]{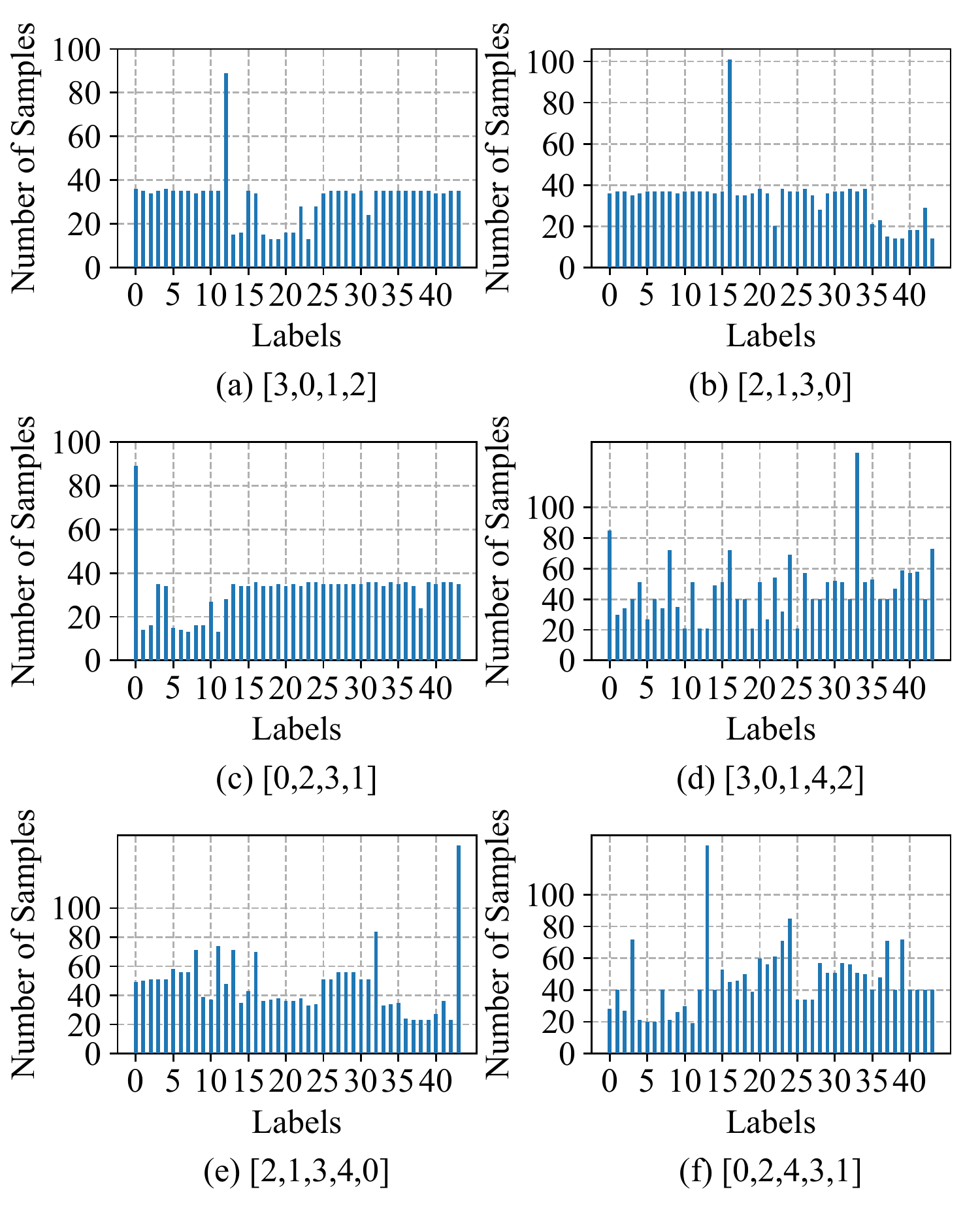}
  \end{subfigure}
  \caption{(a),~(b) and~(c) show the class distribution in memory after training on MSCOCO with different task orders.
    (d),~(e) and (f) show the class distribution in memory after training on NUS-WIDE with different task orders.}
  \label{fig:task-order-memory}
\end{figure}

\begin{figure}[t]
  \centering
  \begin{subfigure}{1\linewidth}
    \includegraphics[width=1.0\linewidth]{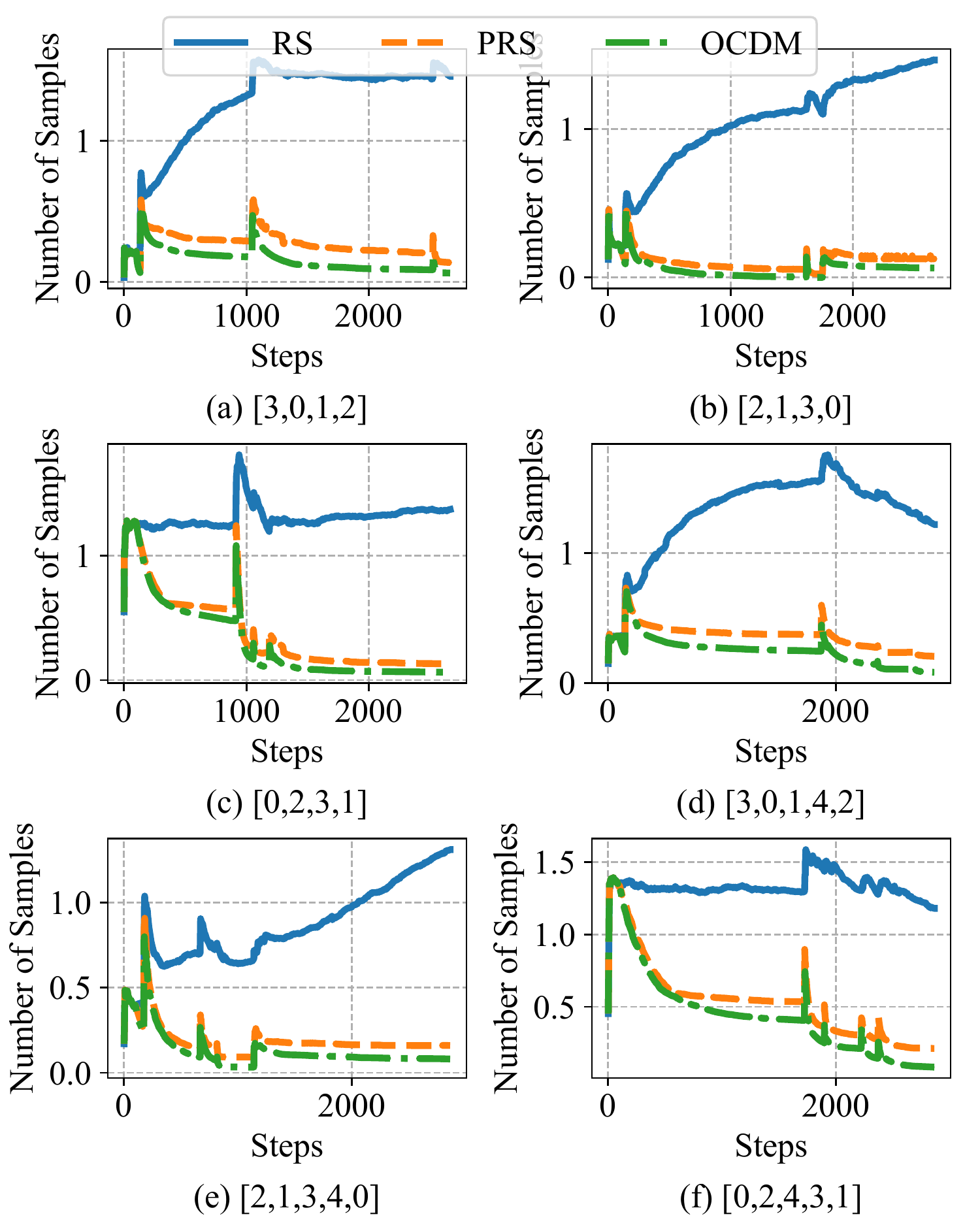}
  \end{subfigure}
  \caption{(a),~(b) and~(c) show the change of distance between class distribution in memory and target distribution with different task orders on MSCOCO.
    (d),~(e) and (f) show the change of distance between class distribution in memory and target distribution with different task orders on NUS-WIDE.}
  \label{fig:task-order-dist}
\end{figure}

Existing works~\cite{masana2020class,ramasesh2020anatomy,yoon2019scalable} claim that a good continual learning algorithm should be robust to different task orders. Therefore, we randomize the task order in MSCOCO and NUS-WIDE to show its influence on our method OCDM and baselines. Specifically, assume the task order we give in the existing experiments are $[0,1,2,3]$ and $[0,1,2,3,4]$ for MSCOCO and NUS-WIDE, respectively. Table~\ref{tab:task-order} gives the mAP results for 3 different task orders. We can find that OCDM consistently gets the best results in all the task orders. Figure~\ref{fig:task-order-memory} gives the class distribution in memory when training on MSCOCO with our OCDM is over, like that in Figure~\ref{memory distribution-coco} but with different task orders. We can find that OCDM can still make the class distribution in memory balanced. Figure~\ref{fig:task-order-dist} gives the change of distance between class distribution in memory and target distribution during the training on MSCOCO like that in Figure~\ref{performance_change}~(c) and~(d) but with different task orders. We can find that OCDM makes the class distribution in memory always converge to the target distribution. 

\begin{table}[t]
  \scriptsize
  \renewcommand\arraystretch{1.3}
  \renewcommand\tabcolsep{7pt}
  \setlength{\belowcaptionskip}{0.1cm}
  \caption{mAP results of the continual learning methods on MSCOCO and NUS-WIDE with different task orders.}
  \label{tab:task-order}
  \centering
  \begin{tabular}{lccc}
    \toprule
    & \multicolumn{3}{c}{MSCOCO}\\
    \cmidrule(r){2-4}
    Methods     & $[3,0,1,2]$ & $[2,1,3,0]$ & $[0,2,3,1]$ \\
    \midrule  
    RS          & 40.6 $\pm$ 0.5& 42.1 $\pm$ 0.7& 39.1 $\pm$ 0.8  \\
    PRS         & 41.0 $\pm$ 0.4 & 44.9 $\pm$ 2.5 & 44.0 $\pm$ 0.6  \\
    OCDM        & \textbf{44.5 $\pm$ 0.9} & \textbf{45.9 $\pm$ 2.5} & \textbf{46.3 $\pm$ 2.1} \\
    \bottomrule
    & \multicolumn{3}{c}{NUS-WIDE}\\
    \cmidrule(r){2-4}
    Methods     & $[3,0,1,4,2]$ & $[2,1,3,4,0]$ & $[0,2,4,3,1]$ \\
    \midrule  
    RS          & 16.3 $\pm$ 0.7 & 18.1 $\pm$ 0.2 & 17.1 $\pm$ 1.0 \\
    PRS         & 19.5 $\pm$ 0.3 & 20.3 $\pm$ 0.7 & 20.5 $\pm$ 0.3 \\
    OCDM        & \textbf{20.4 $\pm$ 0.5} & \textbf{21.1 $\pm$ 0.5} & \textbf{22.2 $\pm$ 0.1} \\
    \bottomrule
  \end{tabular}
\end{table}


\section{Conclusions}
In this paper, we propose a novel replay-based method, called OCDM, for online multi-label continual learning problems where both task identities and task boundaries are not available.
We try to control the class distribution in memory to overcome the imbalanced data stream. 
Experimental results on real multi-label datasets show that OCDM can well control the class distribution in memory and outperform other methods to achieve state-of-the-art performance.

\bibliographystyle{IEEEtran}
\bibliography{IEEEabrv,paper}

\begin{thebibliography}{10}
\providecommand{\url}[1]{#1}
\csname url@samestyle\endcsname
\providecommand{\newblock}{\relax}
\providecommand{\bibinfo}[2]{#2}
\providecommand{\BIBentrySTDinterwordspacing}{\spaceskip=0pt\relax}
\providecommand{\BIBentryALTinterwordstretchfactor}{4}
\providecommand{\BIBentryALTinterwordspacing}{\spaceskip=\fontdimen2\font plus
\BIBentryALTinterwordstretchfactor\fontdimen3\font minus
  \fontdimen4\font\relax}
\providecommand{\BIBforeignlanguage}[2]{{%
\expandafter\ifx\csname l@#1\endcsname\relax
\typeout{** WARNING: IEEEtran.bst: No hyphenation pattern has been}%
\typeout{** loaded for the language `#1'. Using the pattern for}%
\typeout{** the default language instead.}%
\else
\language=\csname l@#1\endcsname
\fi
#2}}
\providecommand{\BIBdecl}{\relax}
\BIBdecl

\bibitem{parisi2019continual}
G.~I. Parisi, R.~Kemker, J.~L. Part, C.~Kanan, and S.~Wermter, ``Continual
  lifelong learning with neural networks: A review,'' \emph{Neural Networks},
  vol. 113, pp. 54--71, 2019.

\bibitem{goodfellow2013empirical}
I.~J. Goodfellow, M.~Mirza, D.~Xiao, A.~Courville, and Y.~Bengio, ``An
  empirical investigation of catastrophic forgetting in gradient-based neural
  networks,'' \emph{arXiv preprint arXiv:1312.6211}, 2013.

\bibitem{zheng2021evolving}
X.~Zheng, Y.~Zhang, S.~Hong, H.~Li, L.~Tang, Y.~Xiong, J.~Zhou, Y.~Wang,
  X.~Sun, P.~Zhu \emph{et~al.}, ``Evolving fully automated machine learning via
  life-long knowledge anchors,'' \emph{IEEE Transactions on Pattern Analysis
  and Machine Intelligence}, vol.~43, pp. 3091--3107, 2021.

\bibitem{hou2019learning}
S.~Hou, X.~Pan, C.~C. Loy, Z.~Wang, and D.~Lin, ``Learning a unified classifier
  incrementally via rebalancing,'' in \emph{IEEE Conference on Computer Vision
  and Pattern Recognition}, 2019, pp. 831--839.

\bibitem{rajasegaran2020itaml}
J.~Rajasegaran, S.~Khan, M.~Hayat, F.~S. Khan, and M.~Shah, ``itaml: An
  incremental task-agnostic meta-learning approach,'' in \emph{IEEE Conference
  on Computer Vision and Pattern Recognition}, 2020, pp. 13\,588--13\,597.

\bibitem{zhu2021prototype}
F.~Zhu, X.-Y. Zhang, C.~Wang, F.~Yin, and C.-L. Liu, ``Prototype augmentation
  and self-supervision for incremental learning,'' in \emph{IEEE Conference on
  Computer Vision and Pattern Recognition}, 2021, pp. 5871--5880.

\bibitem{lopez2017gradient}
D.~Lopez-Paz and M.~Ranzato, ``Gradient episodic memory for continual
  learning,'' in \emph{Advances in Neural Information Processing Systems},
  2017, pp. 6470--6479.

\bibitem{aljundi2019task}
R.~Aljundi, K.~Kelchtermans, and T.~Tuytelaars, ``Task-free continual
  learning,'' in \emph{IEEE Conference on Computer Vision and Pattern
  Recognition}, 2019, pp. 11\,254--11\,263.

\bibitem{aljundi2019gradient}
R.~Aljundi, M.~Lin, B.~Goujaud, and Y.~Bengio, ``Gradient based sample
  selection for online continual learning,'' in \emph{Advances in Neural
  Information Processing Systems}, 2019, pp. 11\,816--11\,825.

\bibitem{chrysakis2020online}
A.~Chrysakis and M.-F. Moens, ``Online continual learning from imbalanced
  data,'' in \emph{International Conference on Machine Learning}, 2020, pp.
  1952--1961.

\bibitem{DBLP:conf/iclr/TangM21}
B.~Tang and D.~S. Matteson, ``Graph-based continual learning,'' in
  \emph{International Conference on Learning Representations}, 2021.

\bibitem{kirkpatrick2017overcoming}
J.~Kirkpatrick, R.~Pascanu, N.~Rabinowitz, J.~Veness, G.~Desjardins, A.~A.
  Rusu, K.~Milan, J.~Quan, T.~Ramalho, A.~Grabska-Barwinska \emph{et~al.},
  ``Overcoming catastrophic forgetting in neural networks,'' \emph{Proceedings
  of the National Academy of Sciences}, vol. 114, pp. 3521--3526, 2017.

\bibitem{zenke2017continual}
F.~Zenke, B.~Poole, and S.~Ganguli, ``Continual learning through synaptic
  intelligence,'' in \emph{International Conference on Machine Learning}, 2017,
  pp. 3987--3995.

\bibitem{ahn2019uncertainty}
H.~Ahn, S.~Cha, D.~Lee, and T.~Moon, ``Uncertainty-based continual learning
  with adaptive regularization,'' in \emph{Advances in Neural Information
  Processing Systems}, 2019, pp. 4392--4402.

\bibitem{shi2021continual}
Y.~Shi, L.~Yuan, Y.~Chen, and J.~Feng, ``Continual learning via bit-level
  information preserving,'' in \emph{IEEE Conference on Computer Vision and
  Pattern Recognition}, 2021, pp. 16\,674--16\,683.

\bibitem{yoon2017lifelong}
J.~Yoon, E.~Yang, J.~Lee, and S.~J. Hwang, ``Lifelong learning with dynamically
  expandable networks,'' \emph{arXiv preprint arXiv:1708.01547}, 2017.

\bibitem{li2019learn}
X.~Li, Y.~Zhou, T.~Wu, R.~Socher, and C.~Xiong, ``Learn to grow: A continual
  structure learning framework for overcoming catastrophic forgetting,'' in
  \emph{International Conference on Machine Learning}, 2019, pp. 3925--3934.

\bibitem{hung2019compacting}
C.-Y. Hung, C.-H. Tu, C.-E. Wu, C.-H. Chen, Y.-M. Chan, and C.-S. Chen,
  ``Compacting, picking and growing for unforgetting continual learning,''
  \emph{Advances in Neural Information Processing Systems}, pp.
  13\,669--13\,679, 2019.

\bibitem{de2019continual}
M.~De~Lange, R.~Aljundi, M.~Masana, S.~Parisot, X.~Jia, A.~Leonardis,
  G.~Slabaugh, and T.~Tuytelaars, ``A continual learning survey: defying
  forgetting in classification tasks,'' \emph{arXiv preprint arXiv:1909.08383},
  2019.

\bibitem{guo2019improved}
Y.~Guo, M.~Liu, T.~Yang, and T.~Rosing, ``Improved schemes for episodic
  memory-based lifelong learning,'' \emph{arXiv preprint arXiv:1909.11763},
  2019.

\bibitem{kim2020imbalanced}
C.~D. Kim, J.~Jeong, and G.~Kim, ``Imbalanced continual learning with
  partitioning reservoir sampling,'' in \emph{European Conference on Computer
  Vision}, 2020, pp. 411--428.

\bibitem{aljundi2019online}
R.~Aljundi, L.~Caccia, E.~Belilovsky, M.~Caccia, M.~Lin, L.~Charlin, and
  T.~Tuytelaars, ``Online continual learning with maximally interfered
  retrieval,'' \emph{arXiv preprint arXiv:1908.04742}, 2019.

\bibitem{buzzega2020dark}
P.~Buzzega, M.~Boschini, A.~Porrello, D.~Abati, and S.~Calderara, ``Dark
  experience for general continual learning: a strong, simple baseline,''
  \emph{arXiv preprint arXiv:2004.07211}, 2020.

\bibitem{guo2022online}
Y.~Guo, B.~Liu, and D.~Zhao, ``Online continual learning through mutual
  information maximization,'' in \emph{International Conference on Machine
  Learning}.\hskip 1em plus 0.5em minus 0.4em\relax PMLR, 2022, pp. 8109--8126.

\bibitem{caccia2022new}
L.~Caccia, R.~Aljundi, N.~Asadi, T.~Tuytelaars, J.~Pineau, and E.~Belilovsky,
  ``New insights on reducing abrupt representation change in online continual
  learning,'' in \emph{International Conference on Learning Representations},
  2022.

\bibitem{vitter1985random}
J.~S. Vitter, ``Random sampling with a reservoir,'' \emph{ACM Transactions on
  Mathematical Software}, vol.~11, pp. 37--57, 1985.

\bibitem{gupta2020maml}
G.~Gupta, K.~Yadav, and L.~Paull, ``La-maml: Look-ahead meta learning for
  continual learning,'' \emph{arXiv preprint arXiv:2007.13904}, 2020.

\bibitem{DBLP:journals/corr/abs-2204-04763}
S.~Sun, D.~Calandriello, H.~Hu, A.~Li, and M.~K. Titsias,
  ``Information-theoretic online memory selection for continual learning,'' in
  \emph{International Conference on Learning Representations}, 2022.

\bibitem{yoon2021online}
J.~Yoon, D.~Madaan, E.~Yang, and S.~J. Hwang, ``Online coreset selection for
  rehearsal-based continual learning,'' in \emph{International Conference on
  Learning Representations}, 2021.

\bibitem{tiwari2022gcr}
R.~Tiwari, K.~Killamsetty, R.~Iyer, and P.~Shenoy, ``Gcr: Gradient coreset
  based replay buffer selection for continual learning,'' in \emph{Proceedings
  of the IEEE/CVF Conference on Computer Vision and Pattern Recognition}, 2022,
  pp. 99--108.

\bibitem{shim2021online}
D.~Shim, Z.~Mai, J.~Jeong, S.~Sanner, H.~Kim, and J.~Jang, ``Online
  class-incremental continual learning with adversarial shapley value,'' in
  \emph{Proceedings of the AAAI Conference on Artificial Intelligence},
  vol.~35, no.~11, 2021, pp. 9630--9638.

\bibitem{du2022agcn}
K.~Du, F.~Lyu, F.~Hu, L.~Li, W.~Feng, F.~Xu, and Q.~Fu, ``Agcn: Augmented graph
  convolutional network for lifelong multi-label image recognition,''
  \emph{arXiv preprint arXiv:2203.05534}, 2022.

\bibitem{yan2021framework}
S.~Yan, J.~Zhou, J.~Xie, S.~Zhang, and X.~He, ``An em framework for online
  incremental learning of semantic segmentation,'' in \emph{Proceedings of
  International Conference on Multimedia}, 2021, pp. 3052--3060.

\bibitem{DBLP:conf/nips/JinSDR21}
X.~Jin, A.~Sadhu, J.~Du, and X.~Ren, ``Gradient-based editing of memory
  examples for online task-free continual learning,'' in \emph{Advances in
  Neural Information Processing Systems}, 2021, pp. 29\,193--29\,205.

\bibitem{de2021continual}
M.~De~Lange and T.~Tuytelaars, ``Continual prototype evolution: Learning online
  from non-stationary data streams,'' in \emph{IEEE International Conference on
  Computer Vision}, 2021, pp. 8250--8259.

\bibitem{prabhu2020gdumb}
A.~Prabhu, P.~H. Torr, and P.~K. Dokania, ``Gdumb: A simple approach that
  questions our progress in continual learning,'' in \emph{European Conference
  on Computer Vision}, 2020, pp. 524--540.

\bibitem{DBLP:conf/iclr/KangXRYGFK20}
B.~Kang, S.~Xie, M.~Rohrbach, Z.~Yan, A.~Gordo, J.~Feng, and Y.~Kalantidis,
  ``Decoupling representation and classifier for long-tailed recognition,'' in
  \emph{International Conference on Learning Representations}, 2020.

\bibitem{DBLP:conf/eccv/WuH0WL20}
T.~Wu, Q.~Huang, Z.~Liu, Y.~Wang, and D.~Lin, ``Distribution-balanced loss for
  multi-label classification in long-tailed datasets,'' in \emph{European
  Conference on Computer Vision}, 2020, pp. 162--178.

\bibitem{lin2014microsoft}
T.-Y. Lin, M.~Maire, S.~Belongie, J.~Hays, P.~Perona, D.~Ramanan,
  P.~Doll{\'a}r, and C.~L. Zitnick, ``Microsoft coco: Common objects in
  context,'' in \emph{European Conference on Computer Vision}, 2014, pp.
  740--755.

\bibitem{chua2009nus}
T.-S. Chua, J.~Tang, R.~Hong, H.~Li, Z.~Luo, and Y.~Zheng, ``Nus-wide: a
  real-world web image database from national university of singapore,'' in
  \emph{International Conference on Image and Video Retrieval}, 2009, pp. 1--9.

\bibitem{chaudhry2020continual}
A.~Chaudhry, N.~Khan, P.~K. Dokania, and P.~H. Torr, ``Continual learning in
  low-rank orthogonal subspaces,'' \emph{arXiv preprint arXiv:2010.11635},
  2020.

\bibitem{masana2020class}
M.~Masana, B.~Twardowski, and J.~Van~de Weijer, ``On class orderings for
  incremental learning,'' \emph{arXiv preprint arXiv:2007.02145}, 2020.

\bibitem{ramasesh2020anatomy}
V.~V. Ramasesh, E.~Dyer, and M.~Raghu, ``Anatomy of catastrophic forgetting:
  Hidden representations and task semantics,'' in \emph{International
  Conference on Learning Representations}, 2020.

\bibitem{yoon2019scalable}
J.~Yoon, S.~Kim, E.~Yang, and S.~J. Hwang, ``Scalable and order-robust
  continual learning with additive parameter decomposition,'' in
  \emph{International Conference on Learning Representations}, 2020.

\end{thebibliography}

\begin{IEEEbiographynophoto}{Yan-Shuo Liang}
  received the B.Sc. degree in
  mathematics from Nanjing University, China,
  in 2020, where he is currently pursuing the Ph.D.
  degree with the Department of Computer Science and Technology. His research interest is in
  continual~(lifelong) learning.
\end{IEEEbiographynophoto}

\begin{IEEEbiographynophoto}{Wu-Jun Li}
  received the B.Sc. and M.Eng.
  degrees in computer science from Nanjing University, China, and the Ph.D. degree in computer
  science from The Hong Kong University of Science
  and Technology. He started his academic career
  as an Assistant Professor with the Department of
  Computer Science and Engineering, Shanghai Jiao
  Tong University. He then joined Nanjing University,
  where he is currently a Professor with the
  Department of Computer Science and Technology.
  His research interests are in machine learning, big
  data and artificial intelligence.
\end{IEEEbiographynophoto}




\end{document}